\documentclass[final,5p,times]{elsarticle}

\usepackage{amssymb}
\usepackage{amsmath}
\usepackage{array}
\usepackage{booktabs}
\usepackage{multirow}
\usepackage{graphicx}
\usepackage{pifont}
\usepackage{url}
\usepackage{caption}
\usepackage[hidelinks]{hyperref}
\usepackage{microtype}
\usepackage{dblfloatfix}
\usepackage{placeins}
\usepackage{xcolor}

\newif\ifwidetable

\journal{Information Fusion}

\begin{document}

\begin{frontmatter}

\title{CAMF-Det: Closure-Aware Multimodal Fusion for LiDAR-Camera 3D Object Detection on UAV Platforms}

\author[aff1]{Yanze Jiang}
\author[aff1]{Yanfeng Gu}
\author[aff1]{Xian Li\corref{cor1}}

\affiliation[aff1]{organization={School of Electronics and Information Engineering, Harbin Institute of Technology}, city={Harbin 150001}, country={China}}

\cortext[cor1]{Corresponding author: xianli@hit.edu.cn}

\begin{abstract}

Multimodal 3D object detection based on LiDAR and cameras has demonstrated excellent performance in ground-vehicle scenarios, but has not been explored for Unmanned Aerial Vehicle (UAV) platforms. In UAV top-down scenes, frequent ground-object occlusion dominated by tree canopies causes spatially varying and modality-dependent information degradation. Existing multimodal fusion frameworks neither explicitly model such ground-object occlusion nor embed occlusion awareness into the detection pipeline, limiting their performance in occluded UAV scenes. To address these challenges, we propose CAMF-Det, a closure-aware multimodal fusion framework for LiDAR-camera 3D object detection on UAV platforms, which derives dual-modal occlusion intensity through physics-inspired modeling, and embeds them as priors throughout the detection pipeline. First, a dual-modal closure modeling module explicitly constructs occlusion intensity ground truth for both modalities offline via a Beer--Lambert-inspired formulation and building-mask correction. Second, using these ground-truth maps as supervision, a dual-modal prediction network converts the offline modeling results into online occlusion intensity predictions under single-frame inference. Third, both ground-truth and predicted occlusion intensity are injected into data augmentation, feature encoding, multimodal fusion, and detection head, enabling adaptive detection under spatially varying and modality-dependent information degradation. Experiments on two self-built UAV-based multimodal datasets, SI3D-DI and SI3D-DII, demonstrate that CAMF-Det achieves the best performance across all difficulty levels, with hard-level $\mathrm{mAP}_{\mathrm{BEV}}$ improvements of 9.43\% and 4.88\% over the best competing methods, respectively. These results confirm the effectiveness of explicit occlusion prior modeling and exploitation for robust multimodal 3D detection in UAV scenes.
\end{abstract}

\begin{keyword}
Multimodal 3D object detection \sep UAV platform \sep Closure-Aware \sep LiDAR-camera fusion \sep Physics-inspired modeling
\end{keyword}

\end{frontmatter}
\section{Introduction}\label{introduction}

Multimodal three-dimensional (3D) object detection based on LiDAR and cameras is a fundamental task in 3D scene understanding, with broad applications in autonomous driving, intelligent surveillance, and military reconnaissance \cite{ref1,ref2,ref3}. By fusing high-precision 3D structural information from LiDAR with rich color and texture cues from cameras, detectors can overcome the perceptual limitations of individual sensors and have demonstrated notable advantages in complex scenes.

Current multimodal 3D object detectors have been developed predominantly
for ground-vehicle platforms. Existing approaches can be grouped into
input-level fusion \cite{ref4,ref5}, unified-feature fusion
\cite{ref6,ref7}, and feature-projection fusion \cite{ref8,ref9}.
Input-level fusion appends image-derived semantics to raw point clouds,
unified-feature fusion transforms both modalities into a common
representation such as bird's-eye view (BEV), and feature-projection fusion
establishes cross-modal correspondences through sensor geometric
parameters. Each category achieves cross-modal alignment at a different
representational level and has yielded solid performance gains on public
benchmarks. Nevertheless, their fusion mechanisms are tailored to ground-vehicle platforms and cannot be directly transferred to UAV platforms, where the observation geometry, scene layout, and degradation patterns differ fundamentally.

UAV-based 3D object detection leverages the high mobility and wide
coverage of UAV platforms to enable efficient large-area perception, and
has attracted growing interest in search-and-rescue, target
reconnaissance, and inspection tasks \cite{ref10,ref11}. Camera-based approaches \cite{ref12,ref13,ref14} have established initial frameworks and benchmarks for UAV-perspective 3D perception, but their 3D localization accuracy is limited by the inherent uncertainty of monocular depth estimation.
LiDAR-based approaches \cite{ref15,ref16,ref17} have demonstrated the
feasibility of airborne 3D detection across diverse scenarios, but they are mostly designed for specific application settings and do not incorporate visual information. The two modalities offer complementary strengths in 3D geometric accuracy and visual discriminability, yet multimodal 3D object detection on single-UAV platforms remains unexplored.

\begin{figure*}[tbp]
\centering
\captionsetup{justification=justified, singlelinecheck=false}
\includegraphics[width=0.82\textwidth]{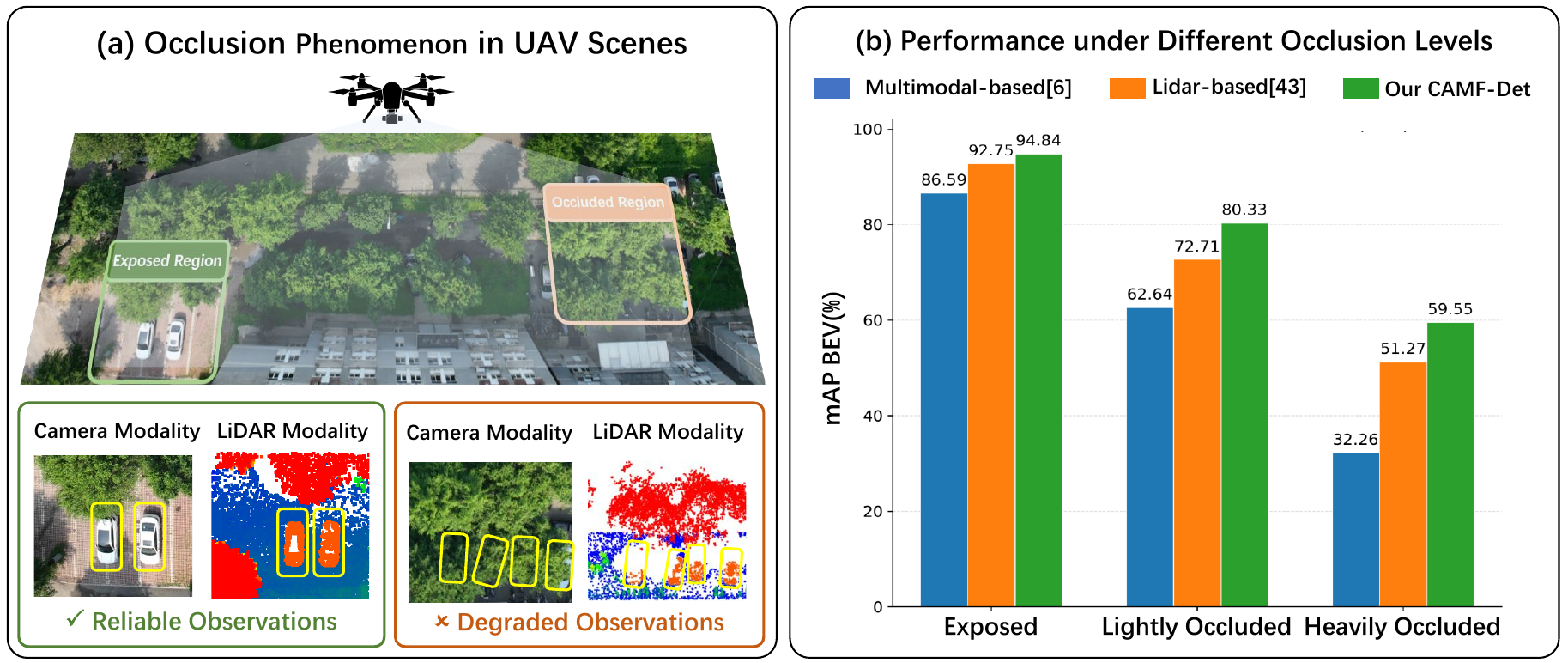}
\caption{Motivation of CAMF-Det. (a) UAV top-down observations suffer from spatially varying and modality-dependent degradation under canopy occlusion. (b) Existing multimodal methods underperform the LiDAR-only baseline, while CAMF-Det achieves the best performance across all occlusion levels.}
\label{fig:1}
\end{figure*}

In UAV top-down scenes, ground targets are frequently occluded by
vegetation canopy, degrading both LiDAR and image observations to
varying degrees across locations. The main reason is that vegetation
canopies consist of randomly distributed branches and leaves that
attenuate both laser pulses and visible light. The resulting information
degradation exhibits pronounced spatial non-uniformity. Moreover, the
two modalities are affected to different degrees. LiDAR pulses can partially penetrate canopy gaps, yielding sparse but still usable point clouds, whereas canopy directly obscures the color and texture cues that the camera relies on, causing more severe image-domain degradation. As a result, observations in open areas remain reliable for both modalities, while those under canopy coverage are degraded to different degrees, as illustrated in Fig.~\ref{fig:1}. Consequently, the main difficulty is
not merely how to fuse two modalities, but how to fuse them under
spatially varying and modality-dependent information degradation. This
gives rise to two key challenges.

First, despite following a well-understood physical mechanism,
canopy-induced attenuation has not been translated into an occlusion
prior that detection networks can directly exploit. As a result,
detectors cannot assess the degree of information degradation across
spatial locations for each modality. Second, existing detection
frameworks lack occlusion-aware designs for UAV scenes. Without
occlusion priors, data augmentation, feature encoding, multimodal
fusion, and detection head are all performed without considering
non-uniform and modality-specific occlusion conditions, making it
difficult to handle the resulting spatially varying and modality-dependent
degradation.

To address both challenges, we propose CAMF-Det, a closure-aware multimodal fusion framework for LiDAR-camera 3D object detection on UAV platforms. The framework consists of three tightly coupled components. To tackle the lack of explicit occlusion modeling, the Dual-Modal Closure Physical Modeling (DPM) module constructs dual-modal occlusion intensity ground truth from multi-strip dense point clouds through Beer–Lambert-inspired closure modeling and building-mask correction. The Dual-Modal Closure Prediction Network (DCPNet) then converts these offline results into online occlusion intensity predictions under single-frame inference. To embed occlusion awareness into the detection pipeline, the Occlusion Prior-Guided Fusion (OPF) strategy injects these priors into data augmentation, feature encoding, multimodal fusion, and detection head, enabling adaptive multimodal perception under spatially varying and modality-dependent degradation.

The main contributions of this paper are as follows:

1) We propose CAMF-Det, to the best of our knowledge the first multimodal fusion framework for LiDAR-camera 3D object detection on a single UAV platform. The main advantage of the framework is its ability to adapt multimodal detection to spatially varying and modality-dependent information degradation.

2) We devise a dual-modal occlusion prior construction and prediction
scheme. The former constructs occlusion intensity ground truth for both modalities offline through Beer--Lambert-inspired closure modeling and building-mask correction, and the latter converts these offline results into online predictions, bridging physical modeling and detection inference.

3) We design an occlusion prior-guided fusion strategy that embeds occlusion intensity into the detection pipeline, exploiting offline ground truth during training and online predictions during both training and inference, enabling adaptive detection in occluded UAV scenes.

The remainder of this paper is organized as follows. Section 2 reviews
related work on Ground-based multimodal and UAV-based 3D object detection. Section 3 presents the proposed CAMF-Det framework and its
core components. Section 4 describes the experimental settings and
reports the results analysis. Section 5 concludes the paper and
discusses future research.

\section{Related works}\label{related-works}

\subsection{Ground-Based LiDAR-Camera 3D Object Detection}

Multimodal 3D object detection based on LiDAR and cameras has become an active
research direction in ground-vehicle perception, as LiDAR provides
accurate 3D geometry while cameras offer rich color and texture cues.
Since LiDAR features usually serve as the geometric foundation of many
fusion frameworks, advances in LiDAR-based detectors have provided
important support for multimodal detection. Existing LiDAR detectors
have evolved from dense \cite{ref18,ref19} and hybrid
\cite{ref20,ref21,ref22} representations to fully sparse architectures
\cite{ref23,ref24,ref25}. Sequence-modeling methods
\cite{ref26,ref27,ref28,ref29} further improve long-range context
modeling. Building on these methods, multimodal detectors mainly differ
in how cross-modal information is integrated, and can be categorized
into input-level fusion, unified-feature fusion, and
feature-projection fusion.

Input-level fusion appends image semantics to raw point clouds during
data preprocessing. Although straightforward to implement, this strategy
is sensitive to calibration errors, and the resulting semantic density
is constrained by point-cloud sparsity. Unified-feature fusion
transforms heterogeneous features from both modalities into a common
geometric space, mainly through BEV-based and virtual-point-based
strategies. BEVFusion \cite{ref6} established a representative BEV
fusion framework, and subsequent work has improved view-transformation
accuracy and mitigated feature misalignment \cite{ref30,ref31,ref32}.
Virtual-point-based methods \cite{ref7,ref33,ref34} convert images into
pseudo point clouds via depth completion, reusing point-cloud detection
pipelines. VirConv \cite{ref34} reduces
virtual-point noise, while CMF-IoU \cite{ref7} improves cross-modal
complementarity mining.

Feature-projection fusion establishes inter-modal correspondences
directly through sensor intrinsic and extrinsic parameters, thereby
achieving cross-modal spatial alignment without explicit depth
estimation. At the encoder stage, AutoAlignV2 \cite{ref35},
VoxelNextFusion \cite{ref36}, and SSLFusion \cite{ref8} improve
alignment through deformable feature sampling, dual-granularity
pixel-and-patch sampling, and multi-stage scale alignment, respectively.
At the detection head stage, TransFusion \cite{ref9} and SparseLIF
\cite{ref37} perform query-driven fusion. The former initializes object
queries from voxel-branch detection results, while the latter generates
query priors from an image detector and introduces uncertainty-aware
adaptive modal weighting.

Although these methods have been extensively validated in ground-vehicle
scenarios, they are not directly applicable to UAV platforms due to
differences in observation geometry and computational constraints. More
fundamentally, they usually assume spatially uniform cross-modal
reliability and overlook canopy-induced information degradation in UAV
top-down scenes, where occlusion varies spatially and affects LiDAR and
image modalities differently. In contrast, our method constructs
physics-inspired dual-modal occlusion priors and exploits them across
multiple detection stages, adapting fusion to spatially varying and
modality-dependent information degradation.

\begin{figure*}[tbp]
\centering
\captionsetup{justification=justified, singlelinecheck=false}
\includegraphics[width=0.9\textwidth]{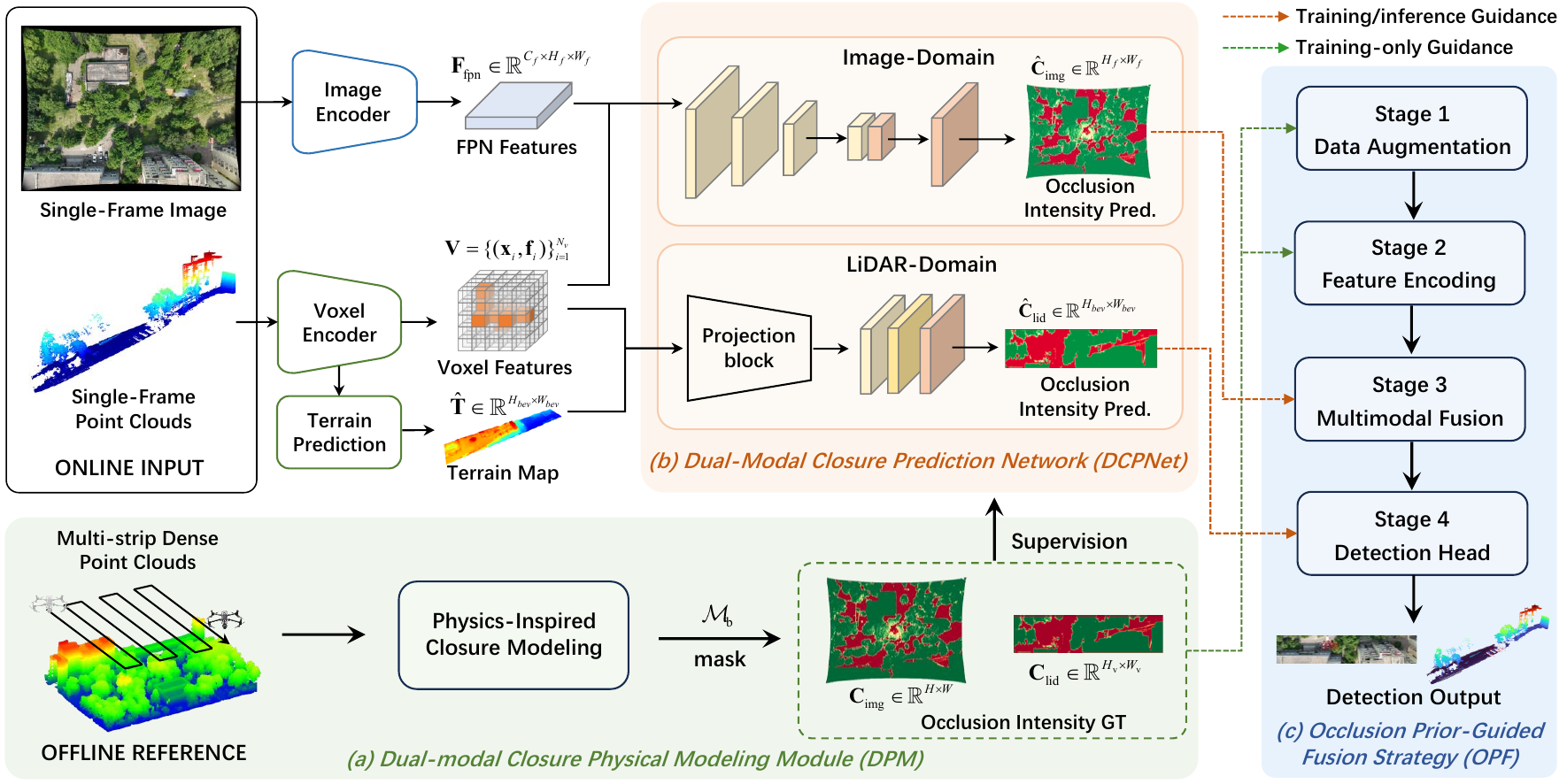}
\caption{Overall architecture of the proposed CAMF-Det. (a) DPM constructs dual-modal occlusion intensity ground truth offline from multi-strip dense point clouds. (b) DCPNet predicts occlusion intensity online by reusing intermediate features from the image and voxel encoders. (c) OPF embeds occlusion priors into detection pipeline. Green and orange arrows distinguish training-only and training/inference guidance, respectively.}
\label{fig:2}
\end{figure*}

\subsection{UAV-Based 3D Object Detection}

Despite the above progress in ground-based multimodal detection, 
directly extending these fusion frameworks to UAV platforms remains 
nontrivial because UAV scenes involve different viewpoints and degradation patterns. In this context, on-board 
LiDAR-camera fusion for a single UAV remains underexplored. Existing 
UAV-based 3D detection work has focused predominantly on single-modal 
methods, with camera-based and LiDAR-based studies developing largely 
independently.

On the camera side, Hu et al.\ \cite{ref38} introduces a geometric deformation transformation module to mitigate
perspective distortion in UAV top-down views, achieving joint 2D and 3D
detection from monocular images. Subsequently, several studies
\cite{ref12,ref14} released UAV-perspective 3D detection datasets using
simulation environments such as CARLA and demonstrated that mainstream
methods degrade considerably under UAV viewpoints. Cooperative
perception efforts have extended the scope from single-UAV to multi-UAV
settings \cite{ref13,ref39} and further to heterogeneous air--ground
collaboration \cite{ref40}. These studies have promoted UAV-perspective
3D perception by providing new benchmarks, detection frameworks, and
collaborative perception paradigms. Nevertheless, camera-based methods
remain fundamentally constrained by the uncertainty of monocular depth
estimation, which limits accurate 3D localization.

On the LiDAR side, existing work has deployed LiDAR sensors on UAVs for
3D detection of traffic objects \cite{ref16,ref17} and human targets
\cite{ref15}, with datasets such as WiSAR3D \cite{ref41} and Pi3DET
\cite{ref42} introduced to support this direction. LiDAR-based methods
benefit from direct 3D geometric measurements and avoid the depth
ambiguity of monocular images. Our prior work TG-ADet \cite{ref43} was
specifically developed for UAV scenarios with complex terrain
distributions and high target--background similarity, and serves as the
single-modal baseline of this work. However, these LiDAR-based efforts
do not incorporate image information, leaving the complementary color
and texture cues of the camera modality unexploited.

In summary, existing UAV-based 3D object detectors have two main limitations. First, current research remains predominantly single-modal or cooperative across multiple platforms. Although recent air–ground cooperative studies \cite{ref44} have introduced LiDAR-camera fusion, their focus is on inter-platform information sharing rather than on-board multimodal detection, and dedicated LiDAR-camera fusion for a single UAV has not been explored. Second, existing methods lack adaptation to the distinctive occlusion-induced degradation in UAV scenes, where canopy occlusion causes spatially varying and modality-dependent information degradation. To address these limitations, we propose CAMF-Det, which explicitly models dual-modal occlusion priors and embeds them across detection pipeline.

\section{Method}\label{method}

This section presents CAMF-Det, whose overall architecture is illustrated in Fig.~\ref{fig:2}. The framework takes a single-frame image and LiDAR point clouds accumulated over a one-second window as inputs, and consists of three coupled components: (a) DPM constructs dual-modal occlusion intensity ground truth from multi-strip dense point clouds; Using these ground-truth maps as supervision, (b) DCPNet predicts the corresponding occlusion intensity online under single-frame inference; (c) OPF embeds occlusion intensity into the detection pipeline, exploiting offline ground truth during training and online predictions during both training and inference.

\subsection{Dual-Modal Closure Physical Modeling}
\begin{figure*}[!t]
\centering
\captionsetup{width=1\textwidth}
\includegraphics[width=0.9\textwidth]{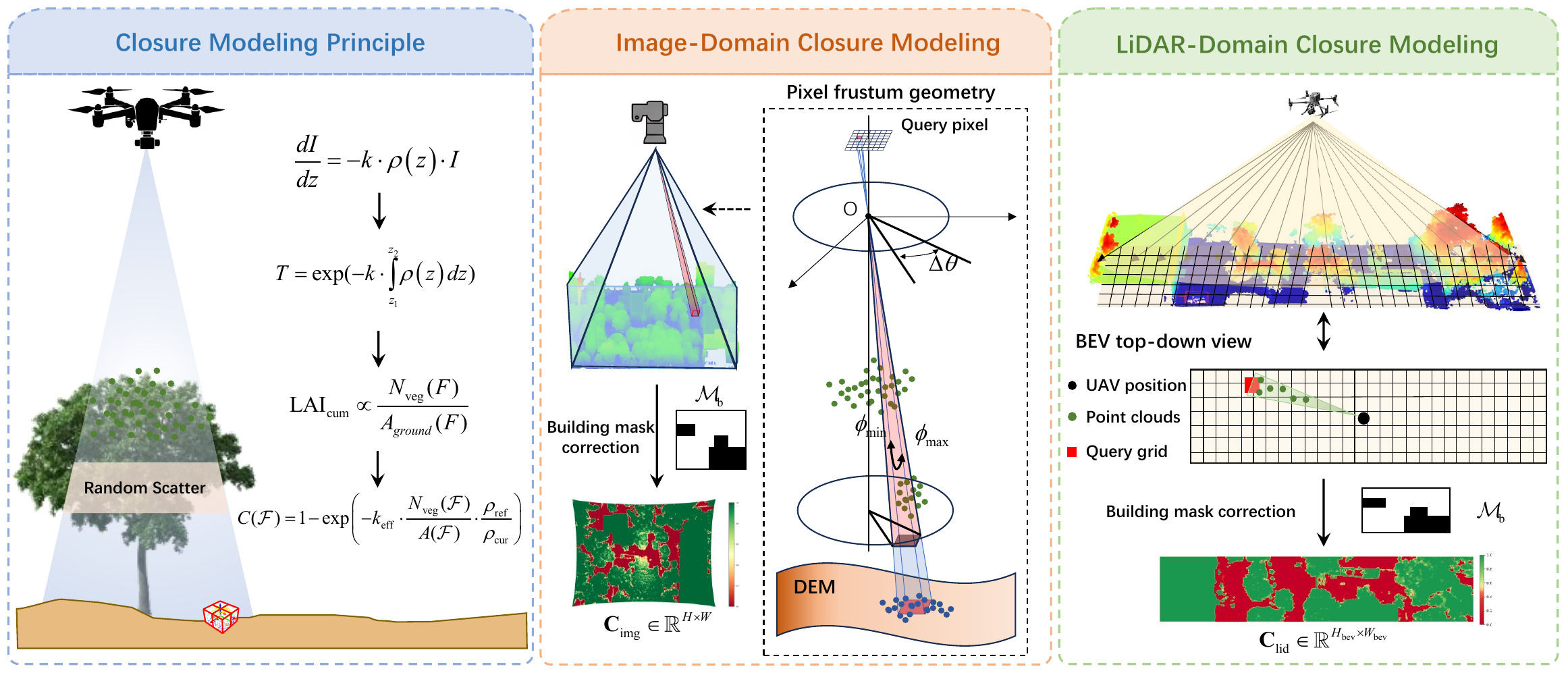}
\caption{Architecture of the DPM module. The Beer–Lambert-inspired closure modeling principle (left) is applied to image-domain pixel frustums (middle) and LiDAR-domain BEV ground grids (right), with building-mask correction, to produce dual-modal occlusion intensity ground truth.}
\label{fig:3}
\end{figure*}

Canopy closure quantifies the degree to which a line of sight is blocked
by vegetation canopy, ranging from 0 (fully open) to 1 (fully
occluded). In this work, \textbf{\emph{closure}} denotes canopy-induced attenuation modeled by the Beer--Lambert law, while \textbf{\emph{occlusion intensity}} further incorporates rigid-body occlusion through building-mask correction. DPM
constructs closure maps for each modality independently via Beer--Lambert-based
modeling. Since single-frame sparse point clouds lack sufficient density for reliable closure estimation, DPM operates on multi-strip dense point clouds that aggregate observations from multiple flight directions, capturing relative spatial variation of canopy attenuation rather than recovering absolute values.
After building-mask correction, the resulting dual-modal occlusion intensity ground truth serves as supervision for DCPNet and as training-stage guidance for OPF, rather than being used during inference. A digital elevation model (DEM) provides terrain
information for ground filtering throughout the pipeline. The overall
pipeline is illustrated in Fig.~\ref{fig:3}.

\subsubsection{Beer--Lambert Closure Modeling}
Tree canopies, composed of randomly distributed branches and leaves, can
be approximated as random scattering media governed by the Beer--Lambert
law \cite{ref45}. For a LiDAR pulse traversing a canopy of thickness $L$
with leaf area density ${{\rho}_{\text{leaf}}}(s)$ at position $s$, the
transmittance is

\begin{equation}
T=\exp \left(
-k\int_{0}^{L}{{\rho}_{\text{leaf}}}(s)ds
\right)=\exp (-k\cdot\text{LA}{{\text{I}}_{\text{cum}}})
\end{equation}
where $k$ is the extinction coefficient and
$\text{LA}{{\text{I}}_{\text{cum}}}=\int_{0}^{L}{{\rho}_{\text{leaf}}}(s)ds$
is the cumulative leaf area index. Closure is the complement of
transmittance.

Since airborne LiDAR cannot directly measure $\rho_{\text{leaf}}$,
$\text{LAI}_{\text{cum}}$ is statistically approximated from multi-strip
dense point clouds \cite{ref46}, which observe the scene from multiple
flight directions to reduce canopy blind spots. For a ground position, a
frustum $\mathcal{F}$ is constructed along the observation direction,
from which the non-ground point count
$N_{\text{veg}}(\mathcal{F})$ and cross-section area
$A(\mathcal{F})$ are computed. Assuming the canopy is
approximately uniform along the observation path, the non-ground point
areal density serves as a statistical proxy for
$\text{LAI}_{\text{cum}}$:

\begin{equation}
\text{LA}{{\text{I}}_{\text{cum}}}={{\rho}_{\text{leaf}}}\cdot
L\propto\frac{{{N}_{\text{veg}}}(\mathcal{F})}{A(\mathcal{F})}
\end{equation}

Substituting into the Beer--Lambert form and introducing an effective
attenuation coefficient $k_{\text{eff}}$ to absorb the unknown
extinction and conversion factors, along with a density normalization
factor $\rho_{\mathrm{ref}}/\rho_{\mathrm{cur}}$ to compensate for
inter-strip sampling differences, the closure surrogate is defined as:

\begin{equation}
\widetilde{C}(\mathcal{F})=1-\exp
\left(
-k_{\mathrm{eff}}\cdot
\frac{N_{\mathrm{veg}}(\mathcal{F})}{A(\mathcal{F})}\cdot
\frac{\rho_{\mathrm{ref}}}{\rho_{\mathrm{cur}}}
\right)
\end{equation}
where $\rho_{\mathrm{cur}}$ is the mean point density of the current
scene and $\rho_{\mathrm{ref}}$ is the dataset-level reference density.
The proposed model is a physics-inspired statistical approximation
rather than a rigorous inversion of canopy optical parameters.

\subsubsection{Image-Domain Closure Modeling}
Image-domain closure modeling assigns a closure value to each pixel,
quantifying canopy occlusion along the camera observation direction. For
pixel $(u,v)$, its four corner points are back-projected into
world-coordinate ray directions using the rectified camera intrinsic
matrix $\mathbf{K}$ and the world-to-camera rotation matrix $\mathbf{R}$,
i.e., ${{\mathbf{d}}_{c}}={{\mathbf{R}}^{\top}}{{\mathbf{K}}^{-1}}
{{[{{u}_{c}},{{v}_{c}},1]}^{\top}}$, $c\in\{1,2,3,4\}$, where
$({{u}_{c}},{{v}_{c}})$ are the image coordinates of the four pixel
corners and ${{\mathbf{d}}_{c}}$ is the corresponding ray direction.

The four directions define the angular extent
$[{{\theta}_{\min}},{{\theta}_{\max}}]\times[{{\phi}_{\min}},{{\phi}_{\max}}]$
of pixel frustum ${{\mathcal{F}}_{u,v}}$ in spherical coordinates, from
which points satisfying the angular constraints are retrieved:

\begin{equation}
\begin{aligned}
\mathcal{P}({{\mathcal{F}}_{u,v}})=\{\mathbf{q}\in\mathcal{P}\mid\;
&\theta(\mathbf{q}-{{\mathbf{p}}_{\text{c}}})\in
[{{\theta}_{\min}},{{\theta}_{\max}}],\\
&\phi(\mathbf{q}-{{\mathbf{p}}_{\text{c}}})\in
[{{\phi}_{\min}},{{\phi}_{\max}}]\}
\end{aligned}
\end{equation}

Ground and non-ground points are separated using the DEM reference
elevation $\bar{h}_{\mathrm{g}}$, yielding
${{N}_{\text{veg}}}({{\mathcal{F}}_{u,v}})$. The frustum cross-section
area is approximated as a spherical patch at the mean ground distance
${{\bar{r}}_{\text{g}}}$, i.e.,
$A({{\mathcal{F}}_{u,v}})={{\bar{r}}_{\text{g}}}^{2}
\Delta\theta |\sin{{\phi}_{\max}}-\sin{{\phi}_{\min}}|$,
where $\Delta\theta={{\theta}_{\max}}-{{\theta}_{\min}}$.
Substituting ${{N}_{\text{veg}}}({{\mathcal{F}}_{u,v}})$ and
$A({{\mathcal{F}}_{u,v}})$ into Eq.(3) yields
$\widetilde{C}({{\mathcal{F}}_{u,v}})$ for each pixel, and traversing
all valid pixels produces the image-domain closure map
${{\mathbf{\widetilde{C}}}_{\text{img}}}\in{{\mathbb{R}}^{H\times W}}$.

\subsubsection{LiDAR-Domain Closure Modeling}
LiDAR-domain closure modeling quantifies canopy occlusion along the
LiDAR observation direction using BEV ground grids as spatial units.
Following the detection coordinate system, the ground plane is
partitioned into
${{H}_{\text{v}}}\times{{W}_{\text{v}}}$ cells at resolution
${{\Delta}_{\text{grid}}}$, with cell elevations obtained from the DEM.
For grid cell $(i,j)$, an observation direction unit vector is defined
from the UAV position $\mathbf{o}$ to the ground position
$\mathbf{g}_{i,j}$ as
${{\widehat{\mathbf{d}}}_{i,j}}=
({{\mathbf{g}}_{i,j}}-\mathbf{o})/
\|{{\mathbf{g}}_{i,j}}-\mathbf{o}\|$.

A narrow cylindrical query volume ${{\mathcal{F}}_{i,j}}$ is
constructed along ${{\widehat{\mathbf{d}}}_{i,j}}$, from which
non-ground points are retrieved and counted as
${{N}_{\text{veg}}}({{\mathcal{F}}_{i,j}})$. Setting
$A({{\mathcal{F}}_{i,j}})=\Delta_{\text{grid}}^{2}$ and substituting
into Eq.\,(3) yields ${{\widetilde{C}}({{\mathcal{F}}_{i,j}})}$ for
each cell, producing the LiDAR-domain closure map
${{\mathbf{\widetilde{C}}}_{\text{lid}}}\in
{{\mathbb{R}}^{{{H}_{\text{v}}}\times{{W}_{\text{v}}}}}$ in BEV
format, spatially aligned with the detection point cloud space.

After obtaining the image- and LiDAR-domain closure maps,
building-covered regions are further corrected because the
Beer--Lambert model applies to random scattering media rather than rigid
occluders. Building point clouds are identified via semantic labels, and
the frustum query framework determines building coverage for each pixel
or grid cell. Morphological closing and hole filling are applied to
compensate for mask incompleteness caused by point cloud sparsity,
yielding the building mask ${{\mathcal{M}}_{\text{b}}}$. Building-covered regions are then corrected as:

\begin{equation}
C(\mathcal{F})=
\begin{cases}
1, & \text{if } \mathcal{F}\in \mathcal{M}_{\mathrm{b}},\\
\widetilde{C}(\mathcal{F}), & \text{otherwise}.
\end{cases}
\end{equation}

The corrected maps ${{\mathbf{C}}_{\text{img}}}$ and
${{\mathbf{C}}_{\text{lid}}}$ constitute the final dual-modal occlusion
intensity ground truth, which jointly represents canopy attenuation and
rigid-object occlusion.

\begin{figure*}[tbp]
\centering
\captionsetup{justification=justified, singlelinecheck=false}
\includegraphics[width=0.73\textwidth]{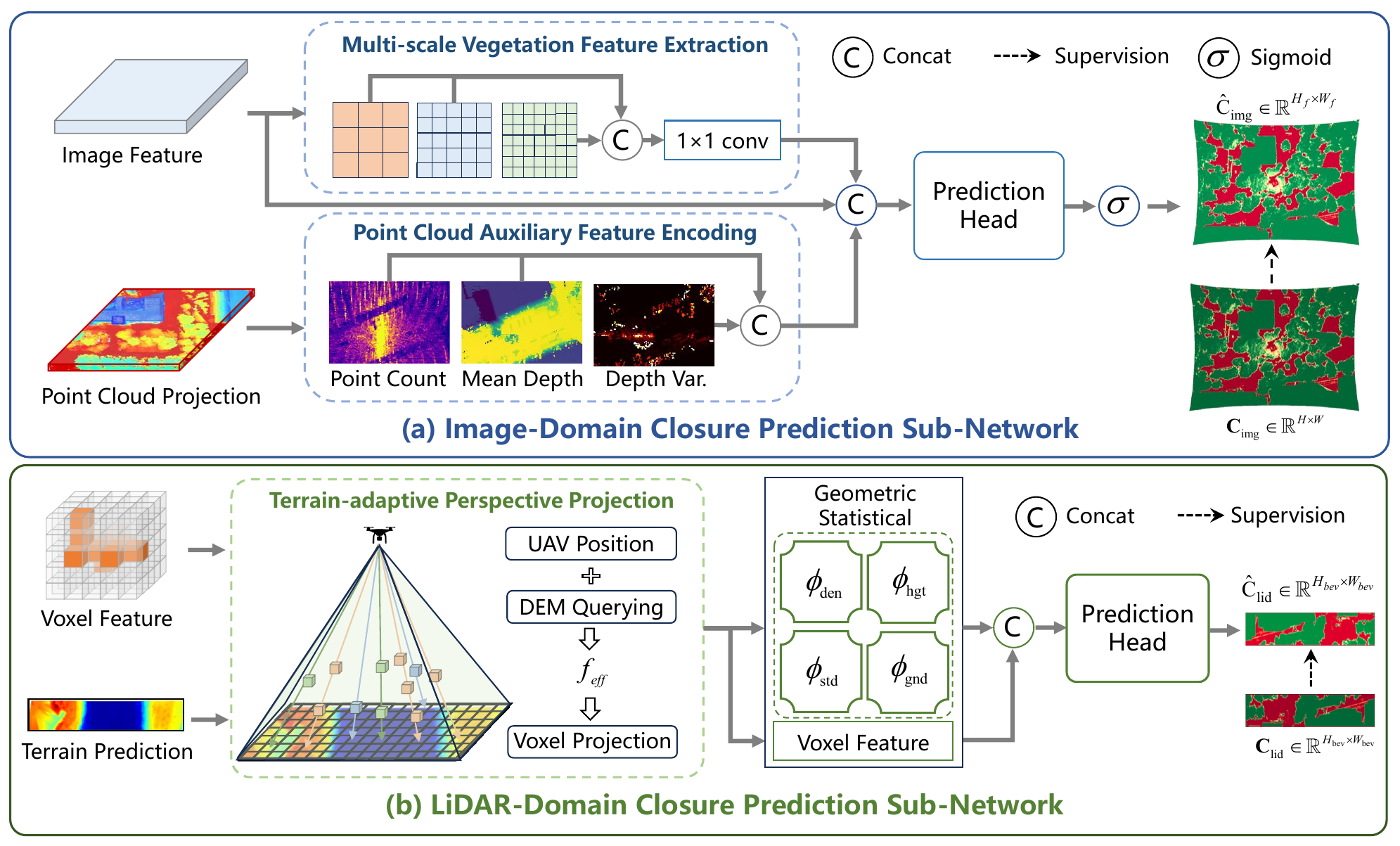}
\caption{Architecture of the DCPNet. (a) Image-domain sub-network with multi-scale vegetation feature extraction and point cloud projection auxiliary encoding. (b) LiDAR-domain sub-network with terrain-adaptive perspective projection and geometric statistical feature extraction.}
\label{fig:4}
\end{figure*}

\subsection{Dual-Modal Closure Prediction Network}
The occlusion intensity ground truth produced by DPM requires
computationally expensive frustum-based point cloud queries and depends
on multi-strip dense point clouds, making it impractical for online
detection. DCPNet is therefore designed to predict the occlusion
intensity distribution of each modality under single-frame inference. To avoid introducing an additional heavy perception branch, it reuses
intermediate features of the detection network. As shown in
Fig.~\ref{fig:4}, the image-domain and LiDAR-domain sub-networks are
structurally independent and are integrated into the image and voxel
branches, respectively, with supervision from the corresponding
DPM ground-truth.

\subsubsection{Image-Domain Closure Prediction Sub-Network}
The image-domain sub-network predicts per-pixel occlusion intensity by combining multi-scale vegetation-sensitive features with sparse point cloud projection cues, enabling the network to perceive canopy boundaries from both visual and geometric perspectives. It takes multi-scale aggregated features
${{\mathbf{F}}_{\text{fpn}}}\in
{{\mathbb{R}}^{{{C}_{f}}\times{{H}_{f}}\times{{W}_{f}}}}$
from the image encoder and feature pyramid, together with point cloud
projection features on the image plane, as inputs. It outputs a
per-pixel occlusion intensity prediction at the feature map resolution
used by the image branch.



A multi-scale vegetation feature extraction module applies three
parallel convolution groups with kernel sizes $3\times 3$, $5\times 5$,
and $7\times 7$ to $\mathbf{F}_{\text{fpn}}$, capturing canopy texture
details, medium-scale crown morphology, and large-scale vegetation
distribution. The multi-scale outputs
$\mathbf{F}_{k}=\operatorname{Conv}_{k\times k}(\mathbf{F}_{\mathrm{fpn}})$,
$k\in\{3,5,7\}$, are concatenated along the channel dimension and fused
via a $1\times 1$ convolution into a vegetation saliency feature map
$\mathbf{F}_{\text{veg}}$, which aggregates canopy-related cues across multiple spatial scales.

Point cloud projections onto the image plane are introduced as geometric
auxiliary cues. The current-frame point cloud is projected onto image
coordinates, and the normalized projected point count, normalized mean
depth, and normalized depth variance at each pixel form a three-channel
projection feature map. This map is encoded through two convolutional
layers into a point cloud auxiliary feature
${{\mathbf{F}}_{\text{pc}}}$ with the same spatial dimensions as
${{\mathbf{F}}_{\text{veg}}}$. These cues help characterize the spatial distribution of returns along each pixel's viewing direction.

The features ${{\mathbf{F}}_{\text{fpn}}}$,
${{\mathbf{F}}_{\text{veg}}}$, and ${{\mathbf{F}}_{\text{pc}}}$ are
concatenated and fused through two convolutional layers. An occlusion
decoder progressively reduces the dimensionality and applies Sigmoid
activation, outputting a single-channel occlusion intensity prediction
${{\widehat{\mathbf{C}}}_{\text{img}}}\in
{{[0,1]}^{{{H}_{f}}\times{{W}_{f}}}}$. The image-domain ground truth
${{\mathbf{C}}_{\text{img}}}$ is used as supervision after bilinear
interpolation for resolution alignment, with the loss computed within
valid mask regions. The loss ${{\mathcal{L}}_{\text{ci}}}$ combines MSE loss,
edge-aware loss, high-closure region consistency loss, and a
smoothness term.

\begin{figure*}[!t]
\centering
\captionsetup{justification=justified, singlelinecheck=false}
\includegraphics[width=0.81\textwidth]{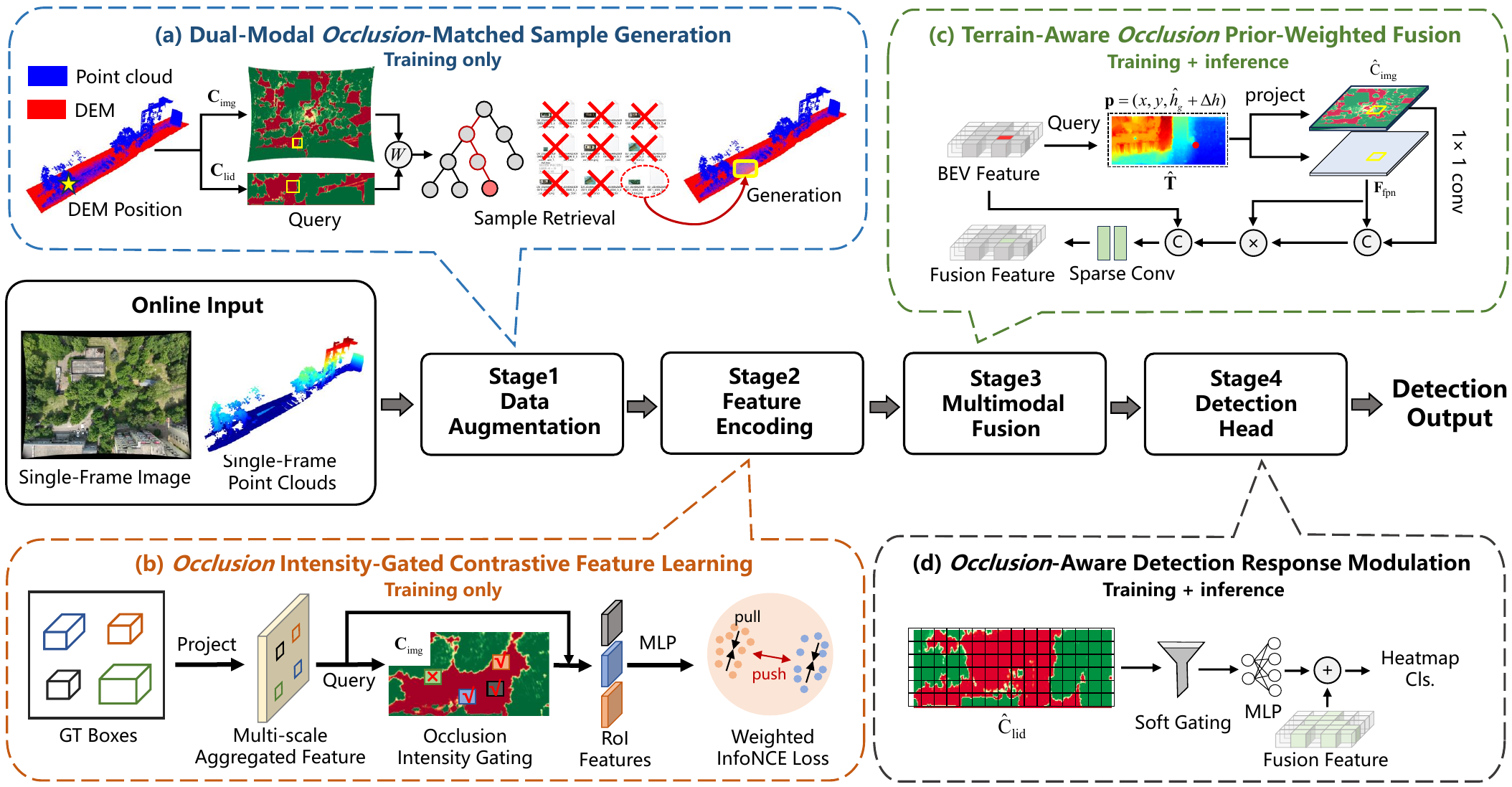}
\caption{Architecture of the OPF strategy, which embeds occlusion priors into data augmentation, feature encoding, multimodal fusion, and detection head stages. Sub-modules (a) and (b) operate during training only, while (c) and (d) are active during both training and inference.}
\label{fig:5}
\end{figure*}

\subsubsection{LiDAR-Domain Closure Prediction Sub-Network}
The LiDAR-domain sub-network estimates BEV occlusion intensity through a terrain-adaptive perspective projection. Unlike DPM, which queries point clouds along frustums from ground grids, this projection starts from existing sparse voxels and maps them onto ground grids defined by predicted terrain elevations, while simultaneously extracting geometric statistical features indicative of local occlusion states. It takes shallow voxel encoder features and a predicted terrain map $\widehat{\mathbf{T}}\in\mathbb{R}^{H_{\mathrm{bev}}\times W_{\mathrm{bev}}}$ as inputs and outputs an occlusion intensity prediction in sparse BEV format. Here, $\widehat{\mathbf{T}}$ is produced by a terrain prediction branch inherited from our prior work TG-ADet \cite{ref30}, which estimates ground elevation across the BEV grid from the input voxels.

Let the sparse voxel features from the first encoder layer be
$\mathbf{V}=\{({{\mathbf{x}}_{i}},{{\mathbf{f}}_{i}})\}_{i=1}^{{{N}_{v}}}$,
where ${{\mathbf{x}}_{i}}=({{X}_{i}},{{Y}_{i}},{{Z}_{i}})$ is the
world coordinate and ${{\mathbf{f}}_{i}}\in{{\mathbb{R}}^{{{C}_{0}}}}$
is the feature vector. The projection layer defines a virtual
perspective model with the UAV position
$\mathbf{o}=({{X}_{\text{uav}}},{{Y}_{\text{uav}}},{{H}_{\text{uav}}})$
as the projection center and the terrain-defined ground grid as the
projection plane. For each voxel $i$, the terrain map
$\widehat{\mathbf{T}}$ is queried at its horizontal location to obtain
the predicted ground elevation $\widehat{h}_{\mathrm{g}}^{i}$, from
which the effective focal length is defined as
$f_{\text{eff}}^{i}=({{H}_{\text{uav}}}-\widehat{h}_{\mathrm{g}}^{i})/
{{r}_{\text{bev}}}$, adaptively correcting the projection for terrain
undulation. The projected coordinates are computed as

\begin{equation}
{{u}_{i}}=f_{\text{eff}}^{i}\cdot
\frac{{{X}_{i}}-{{X}_{\text{uav}}}}{{{H}_{\text{uav}}}-{{Z}_{i}}}+{{u}_{0}},\quad
{{v}_{i}}=f_{\text{eff}}^{i}\cdot
\frac{{{Y}_{i}}-{{Y}_{\text{uav}}}}{{{H}_{\text{uav}}}-{{Z}_{i}}}+{{v}_{0}}
\end{equation}
where $({{u}_{0}},{{v}_{0}})$ denotes the BEV-grid coordinate of the
UAV ground projection. Here, the denominator ${{H}_{\text{uav}}}-{{Z}_{i}}$ determines the perspective scaling along the UAV viewing ray.

For voxels projected onto the same grid cell $(i,j)$, denoted
${{\mathcal{V}}_{i,j}}=\{({{\mathbf{x}}_{m}},{{\mathbf{f}}_{m}})\}_{m=1}^{{{n}_{i,j}}}$,
the projection layer computes a weighted average feature
${{\overline{\mathbf{f}}}_{i,j}}$ and extracts four statistical
features related to canopy occlusion: normalized point density
${{\phi}_{\text{den}}}$, normalized relative ground height
${{\phi}_{\text{hgt}}}$, height standard deviation
${{\phi}_{\text{std}}}$, and ground confidence
${{\phi}_{\text{gnd}}}$. Let $\widehat{h}_{\mathrm{g}}^{i,j}$ be the
predicted terrain elevation of cell $(i,j)$,
$\Delta{{z}_{m}}={{z}_{m}}-\widehat{h}_{\text{g}}^{i,j}$ the relative
height of voxel $m$, and
$\overline{\Delta z}=\sum\nolimits_{m=1}^{{{n}_{i,j}}}{\Delta
{{z}_{m}}}/{{n}_{i,j}}$. The four statistical features are defined as:

\begin{equation}
\begin{alignedat}{2}
\phi_{\mathrm{den}}^{i,j} &=
\frac{n_{i,j}}{n_{\max}}, \qquad&
\phi_{\mathrm{std}}^{i,j} &=
\sqrt{\frac{1}{n_{i,j}}\sum_{m=1}^{n_{i,j}}
(\Delta z_m-\overline{\Delta z})^{2}},\\
\phi_{\mathrm{hgt}}^{i,j} &=
\frac{1}{n_{i,j}}\sum_{m=1}^{n_{i,j}}
\frac{\Delta z_m}{\Delta z_{\max}}, \qquad&
\phi_{\mathrm{gnd}}^{i,j} &=
\mathbb{I}\!\left(\overline{\Delta z}<\tau_h \land
\phi_{\mathrm{std}}^{i,j}<\tau_s\right).
\end{alignedat}
\end{equation}
where ${{n}_{i,j}}$ is the voxel count in cell $(i,j)$,
${{n}_{\max}}$ and $\Delta{{z}_{\max}}$ are the scene-level maximums, 
${{\tau}_{h}}$ and ${{\tau}_{s}}$ are height and standard deviation thresholds (both set to 0.5 by default), and $\mathbb{I}(\cdot)$ is the indicator function. The four features are concatenated with
${{\overline{\mathbf{f}}}_{i,j}}$ and fed into the prediction head.

The prediction head consists of three submanifold sparse convolutional
layers that fuse the projected features while preserving sparsity. The
fused features are compressed into a single channel and clamped to
$[0,1]$, yielding the LiDAR-domain occlusion intensity prediction
${{\widehat{\mathbf{C}}}_{\text{lid}}}$, supervised by
${{\mathbf{C}}_{\text{lid}}}$. The loss
${{\mathcal{L}}_{\text{lid-c}}}$ is the mean of MSE and L1 losses,
with higher weights assigned to heavily occluded regions.

\subsection{Occlusion Prior-Guided Fusion Strategy}
OPF employs four dedicated sub-modules to embed the occlusion intensity ground truth of DPM and the occlusion intensity predictions of DCPNet into the detection pipeline, covering data augmentation, feature encoding, multimodal fusion, and detection head. The internal structure of each sub-module and its correspondence to the detection stages are illustrated in Fig.~\ref{fig:5}. The former two stages exploit DPM ground truth during training only, while the latter two use DCPNet predictions in both training and inference, ensuring that the detector adapts to spatially varying and modality-dependent information degradation.

\subsubsection{Dual-Modal Occlusion-Matched Sample Generation}
GT-Paste \cite{ref20} is widely used in 3D object detection to enrich
training scenes by pasting sample-library targets into new locations.
However, in UAV top-down scenes, canopy occlusion causes target point
cloud density and image visibility to vary significantly across
locations. Standard GT-Paste selects and places samples randomly without
considering whether the occlusion state of a pasted sample is consistent
with the occlusion condition at the placement location, potentially
introducing distribution mismatches between augmented and real targets. 

We propose a dual-modal occlusion-matched sample generation method.
During sample library construction, the mean, standard deviation, and
Gaussian-weighted mean of occlusion intensity are extracted from both
the image-domain and LiDAR-domain ground-truth maps at each target
location. These six statistics are weighted by modality-specific
coefficients and concatenated into a dual-modal occlusion state
descriptor, which is indexed by a per-category KD-tree. During
generation, the reference DEM is queried to obtain the ground elevation
at each candidate position for vertical placement. The same dual-modal
features are extracted at the placement location, and the
nearest-neighbor sample in occlusion state is retrieved from the KD-tree
for pasting. After pasting, the image-domain occlusion intensity ground
truth at the corresponding region is updated via maximum-value fusion to
maintain supervision consistency. This module is used only during
training.

\subsubsection{Occlusion Intensity-Gated Contrastive Feature Learning}
UAV scenes frequently contain targets with similar geometry but
different color and texture, which the image modality captures more
effectively than LiDAR. Contrastive learning can strengthen the image
encoder's discriminative capability for such
fine-grained distinctions. However, image features in heavily occluded
regions degrade severely, and incorporating them into contrastive
learning adversely affects feature space optimization. We therefore use
occlusion intensity as a sample quality gate, ensuring the contrastive
loss operates only on samples with reliable image information.

Target bounding boxes are projected onto the image plane, and fixed-size
local features are extracted and compressed through a projection head
into normalized vectors
${{\mathbf{z}}_{i}}\in
{{\mathbb{R}}^{d}}$. The image-domain
occlusion intensity ground truth at each target center is queried, and
only targets below the threshold ${{\tau
}_{\text{con}}}$ are retained. For the gated set,
same-category targets serve as positives and different-category targets
as negatives. A weighted InfoNCE loss is adopted:

\begin{equation}
{{\mathcal{L}}_{\text{con}}}=-\sum\limits_{i\in
\mathcal{S}}{{{w}_{i}}}\log
\frac{\sum\limits_{j\in
P(i)}{\exp }\left(
{{\mathbf{z}}_{i}}\cdot
{{\mathbf{z}}_{j}}/{{\tau
}_{t}}
\right)}{\sum\limits_{k\in
\mathcal{S},k\ne i}{\exp
}\left(
{{\mathbf{z}}_{i}}\cdot
{{\mathbf{z}}_{k}}/{{\tau
}_{t}} \right)}
\end{equation}
where $\mathcal{S}=\{i\mid
\mathbf{C}_{\mathrm{img}}(\mathbf{u}_{i})<\tau_{\mathrm{con}}\}$
is the gated index set, $\mathbf{u}_{i}$ denotes the projected image
coordinate of the target center, $P(i)=\{j\in\mathcal{S}:y_j=y_i,\,j\ne i\}$
is the set of positive samples for anchor $i$, and $\tau_t$ is the
temperature coefficient.

The weight ${{w}_{i}}$ takes category-specific values,
directing the network to prioritize primary discriminative categories
while including others at lower weights to enrich the negative pool.
The loss is used only during training, and the DPM image-domain
occlusion map provides the gating signal.

\subsubsection{Terrain-Aware Occlusion Prior-Weighted Fusion}
Current multimodal fusion methods, exemplified by BEVFusion \cite{ref6}, transform
image features into BEV space through depth distribution prediction. In
UAV top-down scenes, this paradigm faces depth estimation uncertainty,
high computational overhead from large BEV feature maps, and image
feature degradation under heavy occlusion.

We design a terrain-aware occlusion prior-weighted fusion method. Sparse
LiDAR BEV voxels are projected onto the image plane to query image
features, avoiding depth estimation with computation proportional to BEV
sparsity. The predicted terrain map
$\widehat{\mathbf{T}}$ provides
near-ground 3D query coordinates for each BEV voxel, so that the queried
image features correspond to locations where targets are most likely to
exist. The DCPNet image-domain occlusion intensity prediction
dynamically modulates image feature fusion weights, suppressing degraded
information in heavily occluded regions. This predicted prior is used in
both training and inference.

The fusion proceeds in three stages. In query point generation, the
world coordinates $(x,y)$ of each sparse BEV voxel are extracted, the
terrain map $\widehat{\mathbf{T}}$
provides the predicted ground elevation
$\widehat{h}_{\mathrm{g}}$, and a height offset
$\Delta h$ yields the near-ground query coordinate $\mathbf{p}=(x,y,\widehat{h}_{\mathrm{g}}+\Delta h)$. In projection sampling, these coordinates are
projected onto the image plane via camera intrinsic and extrinsic
matrices, and features are sampled by bilinear interpolation. In
occlusion intensity modulation,
${{\widehat{\mathbf{C}}}_{\text{img}}}$
is interpolated to the image feature map resolution and fed together
with the image features into a dynamic weight generation network:

\begin{equation}
\mathbf{W}=\max \left(
{{\omega }_{\min
}}\text{,}\sigma \left(
\operatorname{Conv}_{1\times 1}\left( \left[
\text{Enc}({{\widehat{\mathbf{C}}}_{\text{img}}});{{\mathbf{F}}_{\text{fpn}}}
\right] \right) \right)
\right)
\end{equation}
where $\operatorname{Enc}(\cdot)$ is a two-layer convolutional encoder,
 $\sigma \left( \cdot
\right)$ is the Sigmoid function, and
$\omega_{\min}$ (default 0.05) is a lower bound
that prevents complete suppression of any feature channel.

The resulting weight map modulates image features spatially and
channel-wise, suppressing contributions from heavily occluded regions
while preserving lightly occluded information. The modulated features,
after projection sampling, are concatenated with BEV voxel features and
fused through sparse convolutions, producing sparse BEV fusion features
${{\mathbf{F}}_{\text{fused}}}$
for the detection head.

\subsubsection{Occlusion-Aware Detection Response Modulation}
Targets in heavily occluded regions suffer from BEV feature degradation
due to sparse point clouds, causing the network to produce lower heatmap
responses. We propose an occlusion-aware detection response
modulation strategy to mitigate this effect. The LiDAR-domain occlusion
intensity prediction of DCPNet serves as a conditional prior, from
which the occlusion intensity value
${{\hat{c}}_{i}}$ is sampled for each valid BEV
feature $i$. This predicted prior is used in both training and
inference. A soft gating function selects heavily occluded regions, and a lightweight mapping network then generates auxiliary features added to the heatmap classification branch:

\begin{equation}
\mathbf{F}_{\text{hm}}^{i}=\mathbf{F}_{\text{fused}}^{i}+\alpha
\cdot \psi (\max
(0,{{\hat{c}}_{i}}-{{\tau
}_{g}}))
\end{equation}
where ${{\tau }_{g}}$ is the gating threshold
(default 0.6), $\psi(\cdot)$ is a lightweight two-layer
MLP mapping the gated occlusion prior to a feature residual, and
$\alpha =\tanh (\gamma
)\cdot {{s}_{\max }}$ is the
modulation intensity coefficient, with $\gamma $ a
zero-initialized learnable parameter and ${{s}_{\max
}}$ the upper bound. This design ensures that the modulation branch
begins as an identity mapping, preserving the pretrained feature
distribution and stabilizing early-stage training.

Features with occlusion intensity below the threshold remain unmodified,
while those in heavily occluded regions receive progressively stronger
enhancement. The modulation is applied only to the heatmap
classification branch, leaving the regression branches unmodified. In
this way, the influence of occlusion priors is confined to target
discrimination while interference with geometric regression is avoided.

\begin{table*}[!t]
\centering
\begin{minipage}{0.8\textwidth}
\caption{Statistical summary of the two datasets.}
\label{tab:1}
\centering
\scriptsize
\setlength{\tabcolsep}{4pt}
\renewcommand{\arraystretch}{1.10}
\begin{tabular*}{\linewidth}{@{\extracolsep{\fill}}ccccccccc@{}}
\toprule
\begin{tabular}[c]{@{}c@{}}\textbf{Dataset}\end{tabular} &
\begin{tabular}[c]{@{}c@{}}\textbf{Survey}\\\textbf{Site}\end{tabular} &
\begin{tabular}[c]{@{}c@{}}\textbf{Survey Area}\\\textbf{($\times 10^{3}\ \text{m}^{2}$)}\end{tabular} &
\begin{tabular}[c]{@{}c@{}}\textbf{Category}\end{tabular} &
\begin{tabular}[c]{@{}c@{}}\textbf{Training}\\\textbf{Instances}\end{tabular} &
\begin{tabular}[c]{@{}c@{}}\textbf{Test}\\\textbf{Instances}\end{tabular} &
\begin{tabular}[c]{@{}c@{}}\textbf{Mean}\\\textbf{Point Count}\end{tabular} &
\begin{tabular}[c]{@{}c@{}}\textbf{Mean Dimension}\\\textbf{L$\times$W$\times$H (m)}\end{tabular} &
\begin{tabular}[c]{@{}c@{}}\textbf{Mean}\\\textbf{Occlusion Intensity}\end{tabular} \\
\midrule
\multirow{4}{*}{SI3D-DI} &
\multirow{4}{*}{\begin{tabular}[c]{@{}c@{}}HIT Campus \&\\Jiangbei Central Park\end{tabular}} &
\multirow{4}{*}{442} &
vehicle    & 10,110 & 10,030 & 548 & 4.83$\times$2.06$\times$1.64 & 0.51 \\
& & & art-target & 2,525  & 1,737  & 217 & 1.44$\times$1.26$\times$0.89 & 0.46 \\
& & & cam-target & 3,025  & 1,783  & 217 & 1.39$\times$1.25$\times$0.85 & 0.43 \\
& & & canopy     & 1,169  & 317    & 579 & 3.38$\times$2.36$\times$1.92 & 0.55 \\
\midrule
SI3D-DII &
\begin{tabular}[c]{@{}c@{}}HIT Science Park\end{tabular} &
351 &
vehicle & 5,711 & 1,951 & 420 & 4.74$\times$1.99$\times$1.65 & 0.30 \\
\bottomrule
\end{tabular*}
\end{minipage}
\end{table*}


\begin{figure*}[!t]
\centering
\captionsetup{justification=justified, singlelinecheck=false}
\includegraphics[width=0.8\textwidth]{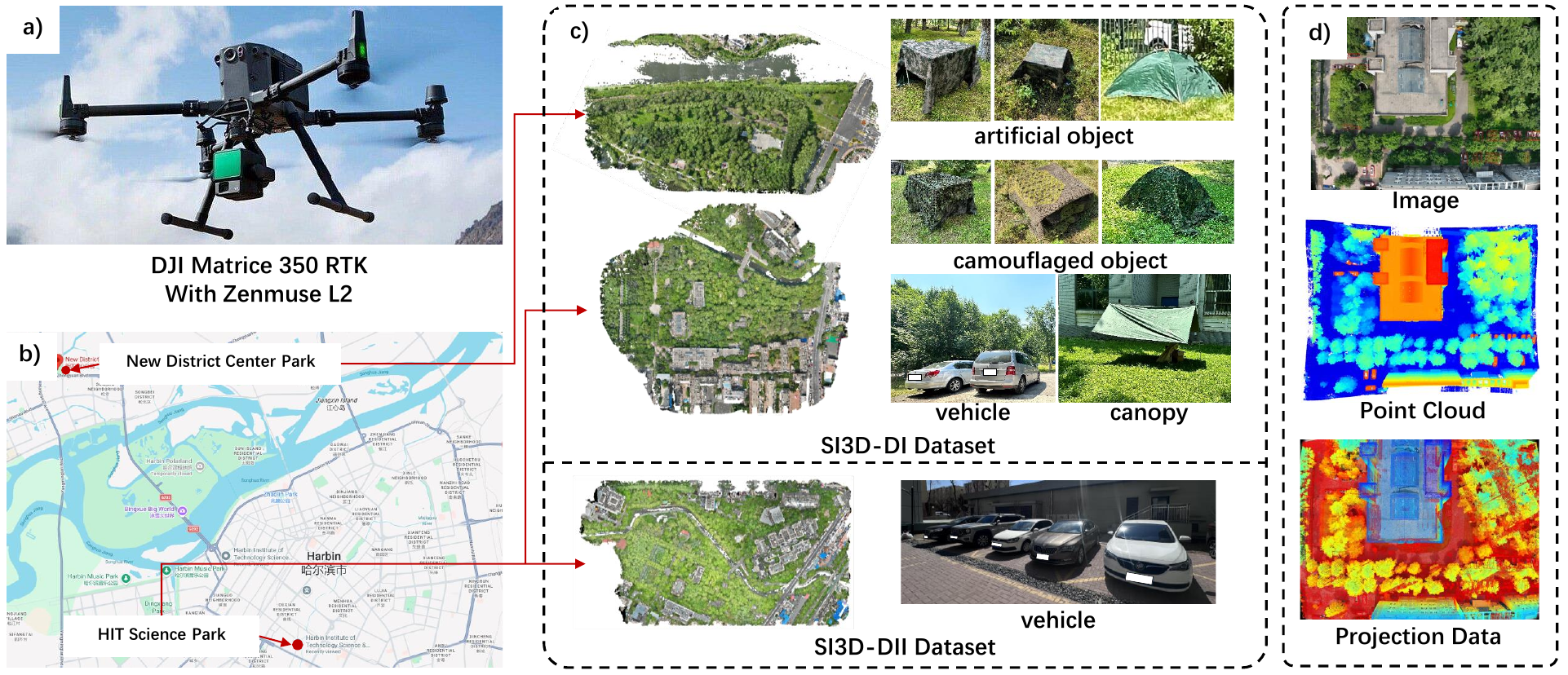}
\caption{Overview of the two UAV-based multimodal datasets. (a) Data acquisition platform. (b) Collection sites. (c) Large-scene point clouds and target categories. (d) Examples of image, point cloud, and LiDAR-to-image projection data illustrating cross-modal spatial alignment in representative scenes.}
\label{fig:6}
\end{figure*}

\begin{figure}[tbp]
\centering
\includegraphics[width=1\linewidth]{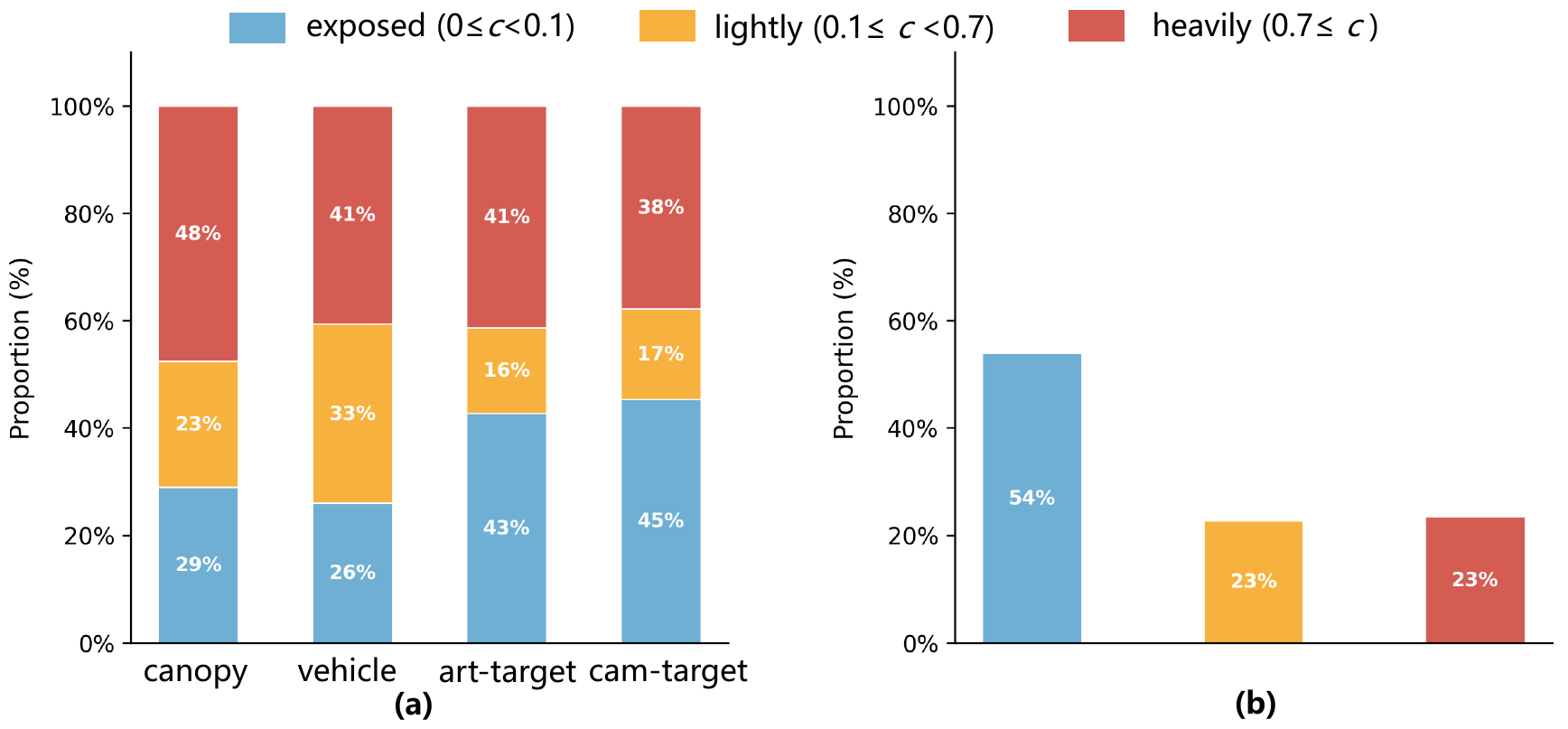}
\caption{Distribution of target instances across exposed, lightly occluded, and heavily occluded levels on the two datasets. (a) SI3D-DI. (b) SI3D-DII.}
\label{fig:7}
\end{figure}

\section{Experimental results and analysis}\label{experimental-results-and-analysis}

\subsection{Experimental datasets}
This work uses two self-built datasets, SI3D-DI and SI3D-DII, both captured by a DJI Matrice 350 RTK UAV equipped with a Zenmuse L2 sensor, which integrates a LiDAR and an optical camera. Sample statistics are reported in Table~\ref{tab:1}, and visualizations of the scenes and targets are shown in Fig.~\ref{fig:6}.

\subsubsection{SI3D-DI Dataset}
This dataset was collected in three sessions from June 26 to 28, 2024,
covering the campus of Harbin Institute of Technology and Jiangbei
Central Park in Harbin, with a total survey area of
$442\times {{10}^{3}}\ {{\text{m}}^{2}}$. The UAV flight altitude was
approximately 70 m, and the LiDAR operated in repetitive scanning mode.
The scene contains naturally distributed vehicles and several manually
deployed targets. After category merging, four detection categories are
defined, namely vehicles, artificial targets (art-target), camouflaged
targets (cam-target), and canopies.

Samples are generated based on image timestamps with a one-second
accumulation window and a 0.2-second frame interval. The training and
test sets are split according to target instance IDs. Samples containing
target instances from the designated test regions are assigned to the
test set, while the remaining samples are used for training. For test
samples that also contain deployed targets from non-test regions, these
instances are masked in both the point cloud and the image to avoid
information leakage. The final dataset contains 4892 training samples
and 3765 test samples, with a total of 30696 instances.

\subsubsection{SI3D-DII Dataset}
This dataset was collected on June 9, 2025, in the Science Park of
Harbin Institute of Technology and its surrounding urban area, covering
$351\times {{10}^{3}}\ {{\text{m}}^{2}}$. The UAV flight altitude was
approximately 60 m, and the LiDAR operated in non-repetitive scanning
mode. The detection targets in this dataset are naturally distributed
vehicles. Samples are generated based on image timestamps with a
one-second accumulation window, with each image corresponding to one
sample. The training--test split follows the same protocol as the
SI3D-DI. The final dataset contains 257 training samples and 102 test
samples, with a total of 7662 vehicle instances.

\widetabletrue
\begin{table*}[!t]
\centering
\caption{Performance of 3D object detection methods on the SI3D-DI test set. The results are reported by the AP with 0.5 IoU threshold. The best and second-best results are highlighted in bold and underlined, respectively.}
\label{tab:2}

\scriptsize
\setlength{\tabcolsep}{1.5pt}
\renewcommand{\arraystretch}{1.08}
\begin{tabular*}{\textwidth}{@{\extracolsep{\fill}}
>{\centering\arraybackslash}p{0.135\textwidth}
>{\centering\arraybackslash}p{0.045\textwidth}
>{\centering\arraybackslash}p{0.070\textwidth}
>{\centering\arraybackslash}p{0.060\textwidth}
*{10}{>{\centering\arraybackslash}p{0.052\textwidth}}
@{}}

\toprule
\multirow{2}{*}{Method} &
\multirow{2}{*}{Modal} &
\multirow{2}{*}{Reference} &
\multirow{2}{*}{Difficult} &
\multicolumn{2}{c}{vehicle} &
\multicolumn{2}{c}{art-target} &
\multicolumn{2}{c}{cam-target} &
\multicolumn{2}{c}{canopy} &
\multicolumn{2}{c}{mAP} \\
\cmidrule(lr){5-6}
\cmidrule(lr){7-8}
\cmidrule(lr){9-10}
\cmidrule(lr){11-12}
\cmidrule(lr){13-14}
& & & &
$\mathrm{AP}_{\mathrm{BEV}}$ & $\mathrm{AP}_{\mathrm{3D}}$ &
$\mathrm{AP}_{\mathrm{BEV}}$ & $\mathrm{AP}_{\mathrm{3D}}$ &
$\mathrm{AP}_{\mathrm{BEV}}$ & $\mathrm{AP}_{\mathrm{3D}}$ &
$\mathrm{AP}_{\mathrm{BEV}}$ & $\mathrm{AP}_{\mathrm{3D}}$ &
$\mathrm{mAP}_{\mathrm{BEV}}$ & $\mathrm{mAP}_{\mathrm{3D}}$ \\
\midrule

\multirow{3}{*}{TransFusion-L \cite{ref9}} &
\multirow{3}{*}{L} &
\multirow{3}{*}{CVPR 2022} &
Easy & 60.43 & 51.59 & 42.24 & 26.59 & 13.92 & 6.36 & 82.74 & 80.43 & 49.83 & 41.24 \\
& & & Moderate & 58.66 & 48.30 & 40.95 & 25.16 & 14.34 & 6.59 & 80.63 & 78.06 & 48.64 & 39.53 \\
& & & Hard & 49.86 & 42.14 & 37.21 & 22.18 & 12.25 & 5.26 & 66.13 & 61.88 & 41.36 & 32.87 \\

\multirow{3}{*}{HEDNet \cite{ref22}} &
\multirow{3}{*}{L} &
\multirow{3}{*}{NeurIPS 2023} &
Easy & 62.77 & 56.37 & 53.44 & \underline{45.85} & 42.29 & 28.50 & 75.86 & 72.62 & 58.59 & 50.83 \\
& & & Moderate & 60.68 & 54.28 & 51.26 & \underline{42.32} & 40.58 & 26.66 & 75.13 & 70.22 & 56.92 & 48.37 \\
& & & Hard & 55.93 & 49.56 & 44.97 & \underline{36.38} & 36.85 & 23.46 & 62.64 & 59.34 & 50.10 & 42.18 \\

\multirow{3}{*}{VoxelNext \cite{ref23}} &
\multirow{3}{*}{L} &
\multirow{3}{*}{CVPR 2023} &
Easy & 74.49 & 67.54 & 47.35 & 33.16 & \underline{46.71} & \underline{31.64} & 74.84 & 70.68 & 60.85 & 50.75 \\
& & & Moderate & 72.26 & 65.27 & 47.02 & 32.69 & \underline{45.57} & \underline{30.36} & 73.26 & 70.33 & 59.53 & 49.66 \\
& & & Hard & 67.23 & 58.36 & 43.38 & 29.62 & \underline{40.81} & \underline{26.39} & 61.43 & 57.49 & 53.21 & 42.97 \\

\multirow{3}{*}{Voxelmamba \cite{ref29}} &
\multirow{3}{*}{L} &
\multirow{3}{*}{NeurIPS 2024} &
Easy & 65.89 & 62.06 & 50.10 & 42.55 & 40.36 & 27.43 & 82.30 & 75.70 & 59.66 & 51.94 \\
& & & Moderate & 63.54 & 59.69 & 45.71 & 39.62 & 38.27 & 25.42 & 79.36 & 71.17 & 56.72 & 48.97 \\
& & & Hard & 56.43 & 50.59 & 38.95 & 33.23 & 33.94 & 21.85 & 64.55 & 58.67 & 48.47 & 41.09 \\

\multirow{3}{*}{SAFDNet \cite{ref24}} &
\multirow{3}{*}{L} &
\multirow{3}{*}{CVPR 2024} &
Easy & 70.57 & 65.54 & 52.23 & 35.21 & 43.17 & 27.55 & 83.83 & \underline{82.13} & 62.45 & 52.61 \\
& & & Moderate & 68.50 & 63.61 & 50.37 & 33.15 & 41.59 & 25.80 & 79.67 & 77.73 & 60.03 & 50.07 \\
& & & Hard & 61.52 & 56.91 & 44.07 & 29.09 & 37.54 & 23.42 & 66.86 & \underline{64.94} & 52.50 & 43.59 \\

\multirow{3}{*}{TG-ADet \cite{ref43}} &
\multirow{3}{*}{L} &
\multirow{3}{*}{TGRS 2025} &
Easy & \underline{74.63} & \textbf{68.09} & \underline{55.88} & 42.73 & 45.46 & 27.47 & 77.53 & 75.53 & \underline{63.37} & \underline{53.45} \\
& & & Moderate & \underline{72.43} & \textbf{65.91} & \underline{53.64} & 40.09 & 44.44 & 27.15 & 75.02 & 72.98 & \underline{61.38} & \underline{51.53} \\
& & & Hard & \underline{67.42} & \underline{60.82} & \underline{47.41} & 35.34 & 39.50 & 24.36 & 63.42 & 61.27 & \underline{54.44} & \underline{45.45} \\
\midrule

\multirow{3}{*}{AutoAlignv2 \cite{ref35}} &
\multirow{3}{*}{L+C} &
\multirow{3}{*}{ECCV 2022} &
Easy & 60.79 & 46.46 & 25.87 & 10.92 & 35.02 & 16.78 & 57.28 & 46.82 & 44.74 & 30.25 \\
& & & Moderate & 59.08 & 44.92 & 25.53 & 10.47 & 34.44 & 15.71 & 57.55 & 46.47 & 44.15 & 29.39 \\
& & & Hard & 52.62 & 39.25 & 22.14 & 8.44 & 29.25 & 13.20 & 46.14 & 37.04 & 37.54 & 24.48 \\

\multirow{3}{*}{BEVFusion \cite{ref6}} &
\multirow{3}{*}{L+C} &
\multirow{3}{*}{ICRA 2023} &
Easy & 60.43 & 51.62 & 42.35 & 26.58 & 13.82 & 6.53 & 83.35 & 80.56 & 49.99 & 41.32 \\
& & & Moderate & 58.67 & 48.31 & 41.07 & 25.06 & 14.18 & 6.69 & 81.24 & \underline{78.43} & 48.79 & 39.62 \\
& & & Hard & 49.85 & 42.17 & 37.23 & 22.11 & 12.16 & 5.30 & 66.55 & 62.04 & 41.45 & 32.91 \\

\multirow{3}{*}{IS-Fusion \cite{ref30}} &
\multirow{3}{*}{L+C} &
\multirow{3}{*}{CVPR 2024} &
Easy & 66.33 & 55.48 & 37.59 & 24.82 & 42.96 & 22.00 & 75.95 & 70.28 & 55.71 & 43.15 \\
& & & Moderate & 62.19 & 53.72 & 36.74 & 23.91 & 42.89 & 22.91 & 72.92 & 67.58 & 53.69 & 42.03 \\
& & & Hard & 55.28 & 47.26 & 31.29 & 19.51 & 38.14 & 19.52 & 56.91 & 52.29 & 45.41 & 34.65 \\

\multirow{3}{*}{SSLFusion \cite{ref8}} &
\multirow{3}{*}{L+C} &
\multirow{3}{*}{AAAI 2025} &
Easy & 66.31 & 57.75 & 23.90 & 20.80 & 19.94 & 11.84 & 34.73 & 34.14 & 36.22 & 31.13 \\
& & & Moderate & 62.45 & 55.51 & 21.83 & 18.77 & 19.48 & 10.84 & 37.87 & 37.25 & 35.41 & 30.59 \\
& & & Hard & 57.47 & 48.81 & 20.20 & 16.68 & 17.38 & 9.84 & 33.33 & 32.69 & 32.09 & 27.01 \\

\multirow{3}{*}{BEVDilation \cite{ref31}} &
\multirow{3}{*}{L+C} &
\multirow{3}{*}{AAAI 2026} &
Easy & 37.29 & 4.32 & 21.87 & 5.84 & 4.11 & 2.14 & \underline{84.58} & 46.33 & 36.96 & 14.66 \\
& & & Moderate & 35.99 & 4.07 & 21.91 & 6.08 & 4.14 & 1.09 & \underline{82.25} & 46.22 & 36.07 & 14.36 \\
& & & Hard & 32.52 & 3.43 & 19.52 & 5.02 & 4.26 & 1.13 & \underline{67.49} & 36.62 & 30.95 & 11.55 \\
\midrule

\multirow{3}{*}{\textbf{CAMF-Det}} &
\multirow{3}{*}{L+C} &
\multirow{3}{*}{Ours} &
Easy & \textbf{74.73} & \underline{67.70} & \textbf{66.80} & \textbf{48.21} & \textbf{55.63} & \textbf{32.21} & \textbf{88.76} & \textbf{86.45} & \textbf{71.48} & \textbf{58.64} \\
& & & Moderate & \textbf{72.65} & \underline{65.75} & \textbf{65.55} & \textbf{46.76} & \textbf{55.78} & \textbf{31.81} & \textbf{86.59} & \textbf{84.05} & \textbf{70.14} & \textbf{57.09} \\
& & & Hard & \textbf{67.86} & \textbf{61.02} & \textbf{60.36} & \textbf{42.28} & \textbf{51.86} & \textbf{28.76} & \textbf{75.38} & \textbf{71.06} & \textbf{63.87} & \textbf{50.78} \\
\bottomrule

\end{tabular*}
\end{table*}
\widetablefalse

\subsection{Experimental Details}
\textbf{Data preprocessing and network configuration.} The point cloud
contains 3D coordinates and reflectance intensity. The voxel size is set
to 0.1 m. The point cloud detection ranges (XYZ axes) are set to
$[-64,-16,0,64,16,16]$ (m) for SI3D-DI and
$[-55,-55,0,55,55,16]$ (m) for SI3D-DII. Raw images are
downsampled by a factor of 2 and input to the network at
$2640\times 1978$ pixels, normalized using standard
ImageNet parameters. CAMF-Det adopts TG-ADet as the LiDAR-modality
detector, and the image branch employs a Swin Transformer pre-trained on
the DOTA dataset. The effective attenuation coefficient
${{k}_{\text{eff}}}$ of the DPM module is set
to 0.001, the ground separation height threshold is set to 1 m, the
spatial resolution of
${{\mathbf{C}}_{\text{lid}}}$
is 0.2 m/pixel, and the resolution of
${{\mathbf{C}}_{\text{img}}}$
matches that of the input image.

\textbf{Training strategy.} All experiments are conducted on two NVIDIA
V100 GPUs using the AdamW optimizer. Training proceeds in three stages.
The single-modal detection network is first trained for 20 epochs. The
detection-related modules are then frozen, and the dual-modal closure
prediction sub-networks are trained independently for 20 epochs.
Finally, the complete fusion network is jointly trained for 10 epochs.
CAMF-Det retains the terrain prediction branch inherited from TG-ADet,
whose output $\widehat{\mathbf{T}}$ is
used by the LiDAR-domain occlusion prediction module and the
terrain-aware fusion module.

\textbf{Evaluation metrics.} Bird's-eye view average
precision
($\mathrm{AP}_{\mathrm{BEV}}$)
and 3D average precision
($\mathrm{AP}_{\mathrm{3D}}$)
are adopted as detection performance metrics, both computed using
40-point interpolation. The IoU threshold for all categories on both
datasets is set to 0.5. The test set is evaluated along two dimensions.
The first dimension stratifies targets into easy, moderate, and hard
difficulty levels based on the number of interior points. The second
dimension stratifies targets into exposed ($0\le
c<0.1$), lightly occluded ($0.1\le
c<0.7$), and heavily occluded ($0.7\le
c\le 1$) levels based on the LiDAR-domain occlusion
intensity in the target neighborhood. The distribution of target
instances across occlusion levels for both test sets is shown in Fig.~\ref{fig:7}.
Occlusion intensity prediction accuracy is evaluated using mean absolute
error (MAE), root mean square error (RMSE), and coefficient of
determination ($R^2$). MAE and RMSE measure the average
deviation and dispersion between predictions and ground truth,
respectively, while $R^2$ reflects the proportion of
ground-truth variance explained by the model.

\subsection{Comparative Analysis With the State-of-the-Art}
To validate the effectiveness of CAMF-Det, we compare it with multiple
representative methods on the SI3D-DI and SI3D-DII datasets. The
compared methods include both LiDAR-based detectors and LiDAR-camera
multimodal detectors. TG-ADet serves as the single-modal baseline in this comparison. The results are presented in Table~\ref{tab:2} and Table~\ref{tab:3},
respectively.

\subsubsection{Performance on the SI3D-DI Test Set}
On the SI3D-DI dataset, CAMF-Det achieves the best results across all
three difficulty levels. Its
$\mathrm{mAP}_{\mathrm{BEV}}$
reaches 71.48\%, 70.14\%, and 63.87\% at the easy, moderate, and hard
levels, improving over TG-ADet by 8.11\%, 8.76\%, and 9.43\%,
respectively. The larger gain at the hard level indicates that the
proposed method is particularly effective for difficult samples.

The performance gains of CAMF-Det vary notably across categories.
Compared with TG-ADet,
$\mathrm{AP}_{\mathrm{BEV}}$
at the easy level improves by 10.92\%, 10.17\%, and 11.23\% for
art-target, cam-target, and canopy, whereas the vehicle category
improves by only 0.10\%. This indicates that the gains are concentrated
on targets more susceptible to occlusion and background interference,
consistent with the design objective of the proposed method.

A notable observation is that most existing multimodal methods perform
poorly on SI3D-DI. For example, BEVFusion achieves only 49.99\%
$\mathrm{mAP}_{\mathrm{BEV}}$
at the easy level, substantially below the 63.37\% of TG-ADet, and other
multimodal methods exhibit a similar trend. Directly introducing image
features does not yield stable improvements in UAV top-down scenes. This is mainly because the spatially varying and modality-dependent nature of occlusion-induced degradation is not taken into account. In contrast, CAMF-Det
leverages explicitly modeled occlusion priors to more effectively
exploit multimodal complementarity.

\begin{figure*}[!t]
\centering
\captionsetup{justification=justified, singlelinecheck=false}
\includegraphics[width=0.84\textwidth]{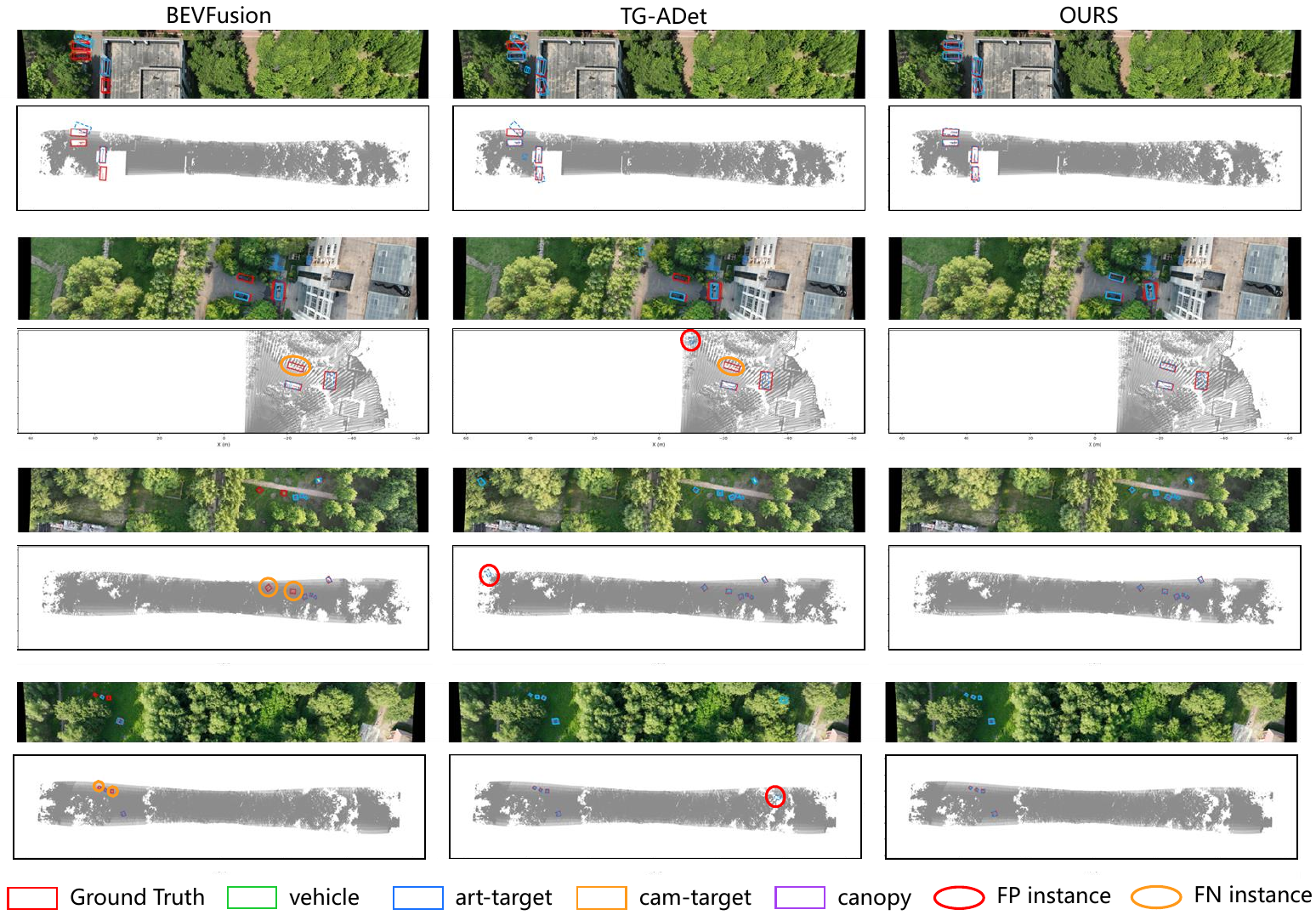}
\caption{Visual comparison of detection results from BEVFusion, TG-ADet, and CAMF-Det on the SI3D-DI dataset. For each scene, the upper and lower rows show the image-domain and LiDAR-domain results, respectively. Circles highlight false positive and false negative instances.}
\label{fig:8}
\end{figure*}

\begin{table*}[!t]
\centering
\caption{Performance of 3D object detection methods on the SI3D-DII test set. The results are reported by the AP with 0.5 IoU threshold. The best and second-best results are highlighted in bold and underlined, respectively.}
\label{tab:3}
\scriptsize
\setlength{\tabcolsep}{4pt}
\renewcommand{\arraystretch}{1.08}
\begin{tabular*}{\textwidth}{@{\extracolsep{\fill}}ccccccccc@{}}
\toprule
\multirow{2}{*}{Method} &
\multirow{2}{*}{Modal} &
\multirow{2}{*}{Reference} &
\multicolumn{3}{c}{$\mathrm{mAP}_{\mathrm{BEV}}$} &
\multicolumn{3}{c}{$\mathrm{mAP}_{\mathrm{3D}}$} \\
\cmidrule(lr){4-6}\cmidrule(lr){7-9}
& & &
Easy & Moderate & Hard &
Easy & Moderate & Hard \\
\midrule
TransFusion-L \cite{ref9} & L & CVPR 2022 & 77.82 & 73.88 & 65.04 & 44.50 & 41.59 & 36.25 \\
HEDNet \cite{ref22} & L & NeurIPS 2023 & 77.67 & 72.81 & 60.69 & 73.70 & 66.63 & 57.00 \\
VoxelNext \cite{ref23} & L & CVPR 2023 & 80.21 & 76.00 & 66.95 & 18.99 & 16.98 & 14.25 \\
Voxelmamba \cite{ref29} & L & NeurIPS 2024 & 37.48 & 33.73 & 28.66 & 18.39 & 16.00 & 13.07 \\
SAFDNet \cite{ref24} & L & CVPR 2024 & \underline{95.13} & 90.32 & 80.44 & 62.96 & 59.32 & 52.94 \\
TG-ADet \cite{ref43} & L & TGRS 2025 & 93.28 & \underline{90.55} & \underline{80.56} & 72.55 & 67.38 & 57.78 \\
\midrule
AutoAlignv2 \cite{ref35} & L+C & ECCV 2022 & 33.55 & 32.87 & 31.22 & 20.54 & 19.70 & 17.64 \\
BEVFusion \cite{ref6} & L+C & ICRA 2023 & 86.77 & 84.46 & 72.83 & \underline{74.13} & \underline{70.07} & \underline{59.26} \\
IS-Fusion \cite{ref30} & L+C & CVPR 2024 & 16.99 & 16.11 & 14.01 & 6.85 & 6.29 & 5.53 \\
SSLFusion \cite{ref8} & L+C & AAAI 2025 & 57.91 & 51.31 & 44.21 & 35.97 & 32.20 & 26.47 \\
BEVDilation \cite{ref31} & L+C & AAAI 2026 & 90.27 & 85.11 & 72.87 & 5.54 & 5.26 & 4.71 \\
\midrule
\textbf{CAMF-Det} & L+C & Ours &
\textbf{95.56} & \textbf{93.01} & \textbf{85.44} &
\textbf{77.04} & \textbf{72.56} & \textbf{63.17} \\
\bottomrule
\end{tabular*}
\end{table*}

\begin{figure*}[tbp]
\centering
\captionsetup{justification=justified, singlelinecheck=false}
\includegraphics[width=0.83\textwidth]{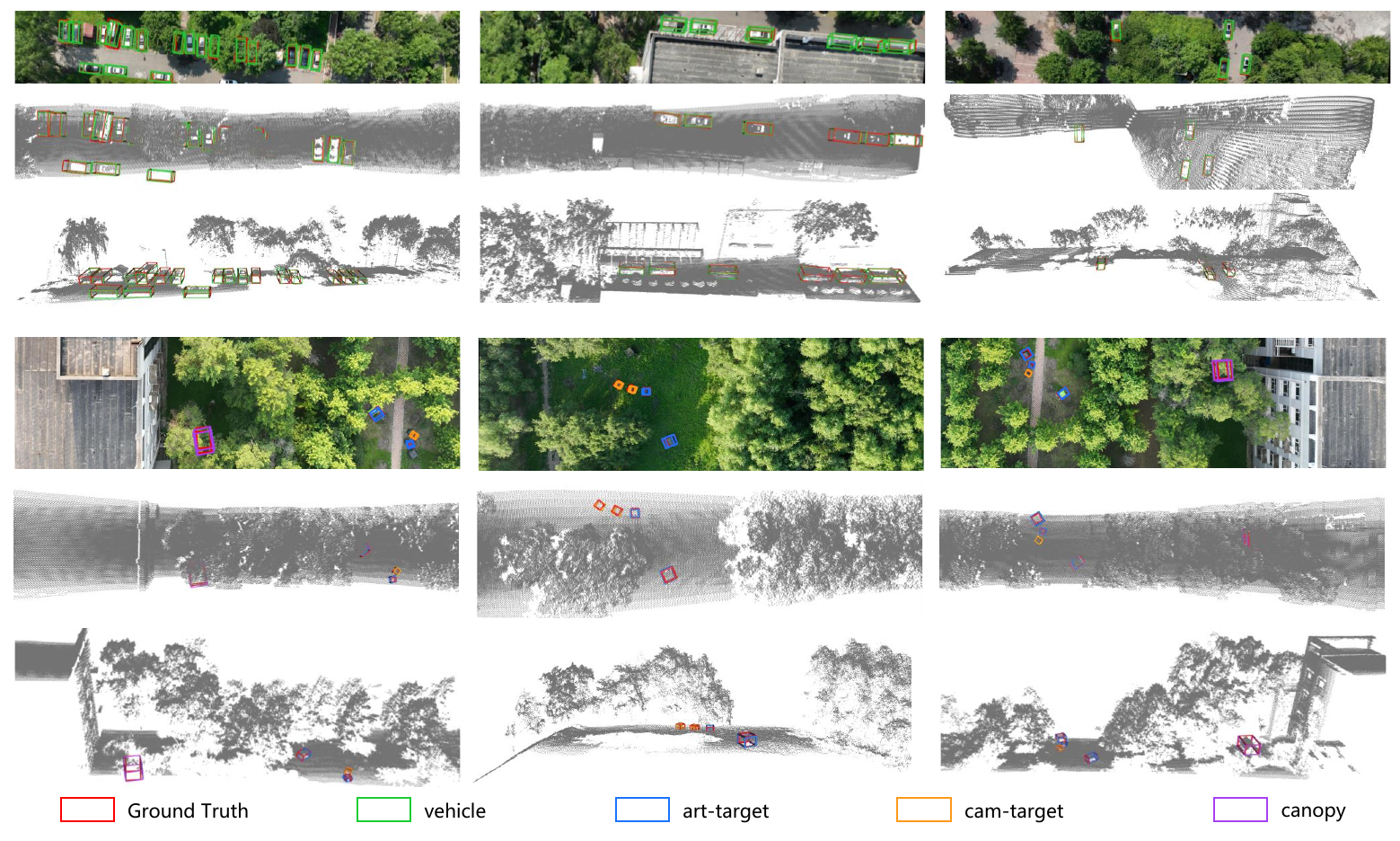}
\caption{Representative detection results of CAMF-Det across six scenes with varying occlusion conditions on the SI3D-DI dataset. For each scene, the first row shows image-domain results, and the second and third rows show LiDAR-domain results from different viewpoints.}
\label{fig:9}
\end{figure*}

Fig.~\ref{fig:8} provides a visual comparison of detection results from BEVFusion,
TG-ADet, and CAMF-Det. CAMF-Det produces more stable detections in
vegetation-occluded regions, densely populated target areas, and complex
backgrounds, with notably fewer false positives and missed detections.
Fig.~\ref{fig:9} further presents the detection results of CAMF-Det in six
representative scenes. These examples show that the proposed method can
produce relatively robust 3D bounding boxes under varying scene
structures and occlusion conditions.

\subsubsection{Performance on the SI3D-DII Test Set}
On the SI3D-DII dataset, CAMF-Det also achieves the best results
across all metrics. Its
$\mathrm{mAP}_{\mathrm{3D}}$
reaches 77.04\%, 72.56\%, and 63.17\% at the easy, moderate, and hard
levels, respectively. Compared with the best existing results,
$\mathrm{mAP}_{\mathrm{3D}}$
improves by 2.91\%, 2.49\%, and 3.91\%. The largest gain is observed at
the hard level, indicating that the proposed method exhibits a more
pronounced advantage on difficult samples.

In contrast to the generally poor performance of multimodal methods on
SI3D-DI, several multimodal methods perform relatively well on
SI3D-DII. For example, BEVFusion achieves
$\mathrm{mAP}_{\mathrm{BEV}}$
of 86.77\%, 84.46\%, and 72.83\%, already surpassing most single-modal
methods. This can be attributed to the fact that SI3D-DII contains
only the vehicle category with a relatively lower overall occlusion
level. Nevertheless, CAMF-Det maintains superior performance, confirming
that the explicit introduction of occlusion priors provides consistent
gains under different scene conditions.

Combining the results on both datasets, CAMF-Det achieves leading
performance across different scanning modes, scene types, and category
configurations. Its advantage is particularly prominent on difficult
categories and hard-level samples, which validates the effectiveness
and generalization capability of the proposed method in UAV scenes.

\begin{table}[thb]
\centering
\captionsetup{justification=justified, singlelinecheck=false}
\caption{Performance comparison under different occlusion levels on the SI3D-DI test set. The best and second-best results are highlighted in bold and underlined, respectively.}
\label{tab:4}
\scriptsize
\setlength{\tabcolsep}{3pt}
\renewcommand{\arraystretch}{1.08}
\begin{tabular*}{\linewidth}{@{\extracolsep{\fill}}cccc@{}}
\toprule
\multirow{2}{*}{Occlusion Level} & \multicolumn{3}{c}{Method} \\
\cmidrule(lr){2-4}
& TG-ADet & Multimodal Baseline & Ours \\
\midrule
Exposed  & 41.05 & \underline{42.74} \textcolor{red}{\scriptsize(+1.69)} & \textbf{46.92} \textcolor{red}{\scriptsize(+4.18)} \\
Lightly  & 42.49 & \underline{49.47} \textcolor{red}{\scriptsize(+6.98)} & \textbf{51.39} \textcolor{red}{\scriptsize(+1.92)} \\
Heavily  & 35.54 & \underline{35.63} \textcolor{red}{\scriptsize(+0.09)} & \textbf{37.14} \textcolor{red}{\scriptsize(+1.51)} \\
\bottomrule
\end{tabular*}
\end{table}


\subsection{Comparative Analysis of Different Occlusion Levels}
To further verify the adaptability of CAMF-Det to occluded scenes, we introduce a multimodal baseline that retains only the basic feature-projection fusion structure with all physics-guided mechanisms removed, thereby isolating the contribution of occlusion prior from that of multimodal fusion itself. We then compare the
$\mathrm{mAP}_{\mathrm{BEV}}$
of CAMF-Det, TG-ADet, and this baseline for targets within different
occlusion intervals on two datasets. The results are presented in Table~\ref{tab:4} and Table~\ref{tab:5}.

\begin{table}[!t]
\centering
\captionsetup{justification=justified, singlelinecheck=false}
\caption{Performance comparison under different occlusion levels on the SI3D-DII test set. The best and second-best results are highlighted in bold and underlined, respectively.}
\label{tab:5}
\scriptsize
\setlength{\tabcolsep}{3pt}
\renewcommand{\arraystretch}{1.08}
\begin{tabular*}{\linewidth}{@{\extracolsep{\fill}}cccc@{}}
\toprule
\multirow{2}{*}{Occlusion Level} & \multicolumn{3}{c}{Method} \\
\cmidrule(lr){2-4}
& TG-ADet & Multimodal Baseline & Ours \\
\midrule
Exposed  & 92.75 & \underline{93.80} \textcolor{red}{\scriptsize(+1.05)} & \textbf{94.84} \textcolor{red}{\scriptsize(+1.04)} \\
Lightly  & 72.71 & \underline{78.40} \textcolor{red}{\scriptsize(+5.69)} & \textbf{80.33} \textcolor{red}{\scriptsize(+1.93)} \\
Heavily  & 51.27 & \underline{51.56} \textcolor{red}{\scriptsize(+0.29)} & \textbf{59.55} \textcolor{red}{\scriptsize(+7.99)} \\
\bottomrule
\end{tabular*}
\end{table}

On the SI3D-DI dataset, CAMF-Det achieves
$\mathrm{mAP}_{\mathrm{BEV}}$
of 46.92\%, 51.39\%, and 37.14\% in the exposed, lightly occluded, and
heavily occluded intervals, obtaining the best results across all
intervals. In the exposed interval, CAMF-Det improves over TG-ADet and
the multimodal baseline by 5.87\% and 4.18\%, respectively. Notably, the multimodal baseline shows some improvement over TG-ADet in the exposed and lightly occluded intervals but provides almost no gain under heavy occlusion. This indicates that basic multimodal fusion
without explicit occlusion modeling fails to stably exploit image
information when occlusion is severe. In contrast, CAMF-Det maintains
superior performance across all three intervals, demonstrating that
occlusion prior guidance effectively mitigates the adverse impact of
image degradation on fusion results.

On the SI3D-DII dataset, CAMF-Det achieves
$\mathrm{mAP}_{\mathrm{BEV}}$
of 94.84\%, 80.33\%, and 59.55\% in the three occlusion intervals, again
obtaining the best results. In the heavily occluded interval, CAMF-Det
improves over TG-ADet and the multimodal baseline by 8.28\% and 7.99\%,
respectively, with margins significantly larger than those in the
exposed and lightly occluded intervals. As the occlusion level
increases, the advantage of CAMF-Det over both baselines widens notably.
This trend indicates that explicit occlusion priors help preserve
reliable cross-modal information under severe occlusion.

Overall, CAMF-Det outperforms both the single-modal and multimodal
baselines across all occlusion intervals on both datasets, with the
gains primarily concentrated on heavily occluded samples. This trend
further validates the effectiveness of the proposed method designed for
occlusion-degraded scenes.

\begin{table}[tbp]
\centering
\captionsetup{justification=justified, singlelinecheck=false}
\caption{LiDAR-domain occlusion intensity prediction accuracy and ablation results of DCPNet on both test sets. Geom.\ stat.\ denotes geometric statistical.}
\label{tab:6}
\fontsize{8pt}{9pt}\selectfont
\setlength{\tabcolsep}{4pt}
\renewcommand{\arraystretch}{1.12}
\begin{tabular*}{\linewidth}{@{\extracolsep{\fill}}ccccc@{}}
\toprule
Network & Dataset & MAE$\downarrow$ & RMSE$\downarrow$ & $R^{2}\uparrow$ \\
\midrule
w/o geom.\ stat.\ features
& \multirow{2}{*}{SI3D-DI} & 0.162 & 0.269 & 0.542 \\
DCPNet 
& & 0.157 & 0.254 & 0.592 \\
\midrule
w/o geom.\ stat.\ features
& \multirow{2}{*}{SI3D-DII} & 0.205 & 0.280 & 0.502 \\
DCPNet 
& & 0.167 & 0.245 & 0.617 \\
\bottomrule
\end{tabular*}
\end{table}

\begin{figure*}[!t]
\centering
\captionsetup{justification=justified, singlelinecheck=false}
\includegraphics[width=0.8\textwidth]{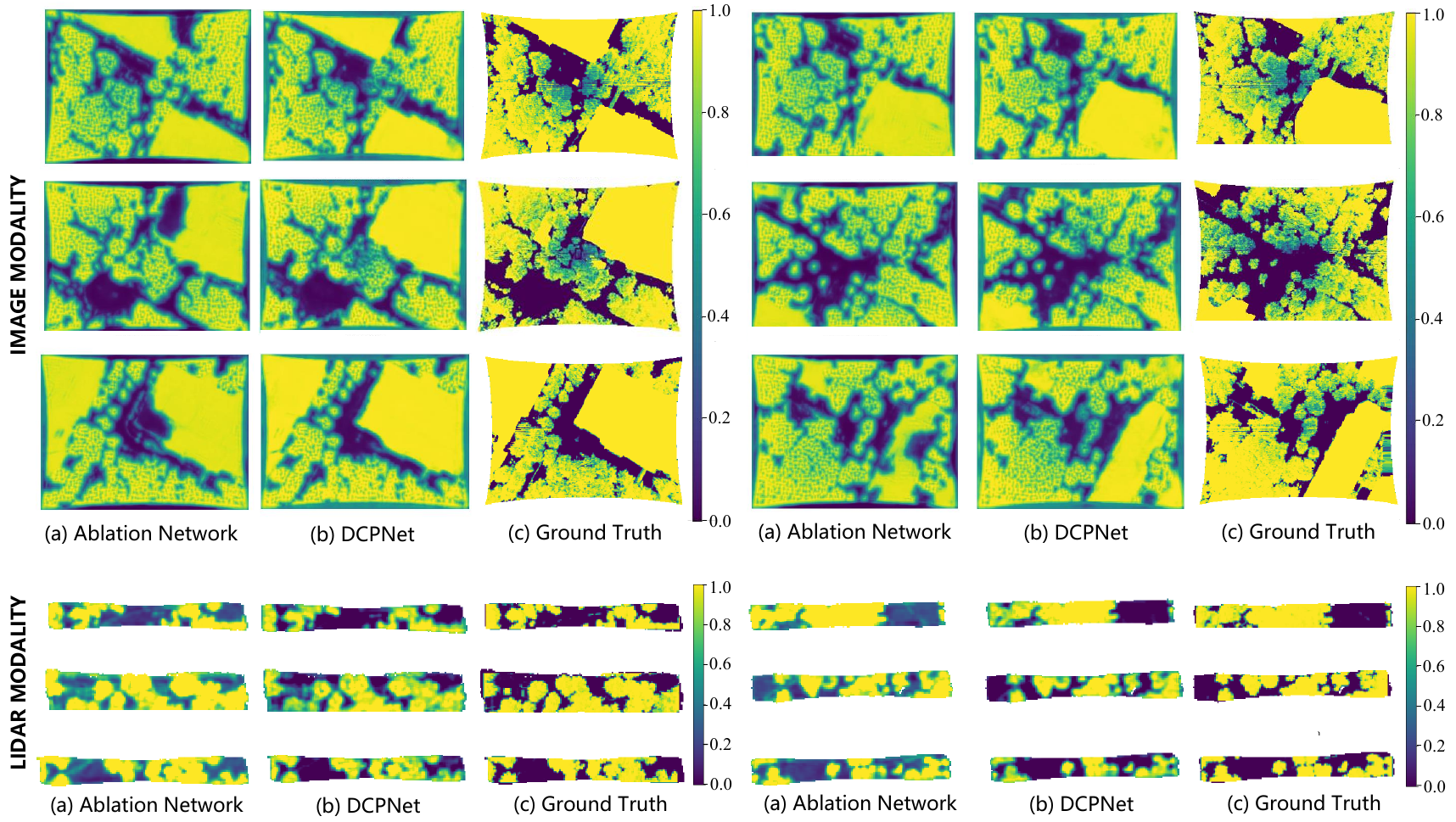}
\caption{Visual comparison of occlusion intensity predictions from the ablation networks, DCPNet, and the corresponding ground truth for both modalities on the test sets. The ablation network removes LiDAR projection features for the image modality and geometric statistical features for the LiDAR modality.}
\label{fig:10}
\end{figure*}

\subsection{Comparative Analysis of DCPNet}
To evaluate the ability of DCPNet to predict the DPM-defined occlusion
intensity ground truth, we compare the prediction results of the two
modal branches and analyze the contribution of auxiliary features
through ablation experiments. For the image modality, the complete
network is compared with a variant that removes the LiDAR point
projection auxiliary feature. For the LiDAR modality, the complete
network is compared with a variant without the geometric statistical
features. The results are presented in Table~\ref{tab:6} and Table~\ref{tab:7}.

\begin{figure}[!t]
\centering
\includegraphics[width=\linewidth]{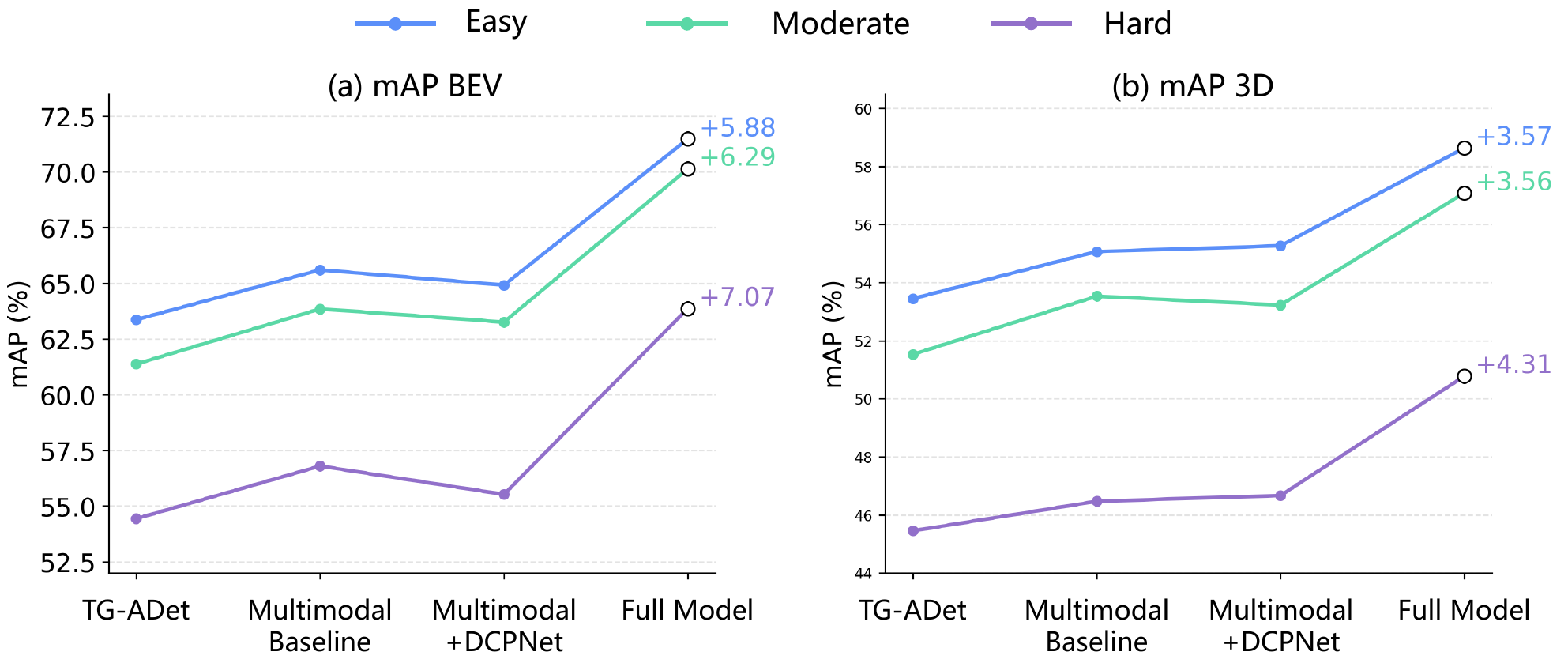}
\captionsetup{justification=justified, singlelinecheck=false}
\caption{Overall framework ablation on SI3D-DI, showing detection performance as the image branch, DCPNet, and OPF are progressively introduced to TG-ADet.}
\label{fig:11}
\end{figure}

\begin{table}[tbp]
\centering
\captionsetup{justification=justified, singlelinecheck=false}
\caption{Image-domain occlusion intensity prediction accuracy and ablation results of DCPNet on both test sets. Proj.\ denotes projection.}
\label{tab:7}
\fontsize{8pt}{9pt}\selectfont
\setlength{\tabcolsep}{4pt}
\renewcommand{\arraystretch}{1.12}
\begin{tabular*}{\linewidth}{@{\extracolsep{\fill}}ccccc@{}}
\toprule
Network & Dataset & MAE$\downarrow$ & RMSE$\downarrow$ & $R^{2}\uparrow$ \\
\midrule
w/o LiDAR proj.\ features
& \multirow{2}{*}{SI3D-DI} & 0.116 & 0.214 & 0.693 \\
DCPNet 
& & 0.111 & 0.210 & 0.704 \\
\midrule
w/o LiDAR proj.\ features
& \multirow{2}{*}{SI3D-DII} & 0.166 & 0.242 & 0.653 \\
DCPNet 
& & 0.146 & 0.220 & 0.714 \\
\bottomrule
\end{tabular*}
\end{table}

\begin{table*}[tbp]
\centering
\begin{minipage}{\textwidth}
\caption{Progressive ablation results of the OPF sub-modules on the SI3D-DI test set.}
\label{tab:8}
\centering
\fontsize{8pt}{9pt}\selectfont
\setlength{\tabcolsep}{3pt}
\renewcommand{\arraystretch}{1.15}
\begin{tabular*}{\linewidth}{@{\extracolsep{\fill}}cccccccccc@{}}
\toprule
\multirow{2}{*}{OPF-Mod} &
\multirow{2}{*}{OPF-Fus} &
\multirow{2}{*}{OPF-Aug} &
\multirow{2}{*}{OPF-Con} &
\multicolumn{3}{c}{$\mathrm{mAP}_{\mathrm{BEV}}$} &
\multicolumn{3}{c}{$\mathrm{mAP}_{\mathrm{3D}}$} \\
\cmidrule(lr){5-7}\cmidrule(lr){8-10}
& & & &
Easy & Moderate & Hard &
Easy & Moderate & Hard \\
\midrule
-- & -- & -- & -- &
63.37 & 61.38 & 54.44 & 53.45 & 51.53 & 45.45 \\
\checkmark & & & &
64.56 \textcolor{red}{\scriptsize(+1.19)} & 63.15 \textcolor{red}{\scriptsize(+1.77)} & 55.94 \textcolor{red}{\scriptsize(+1.50)} & 53.85 \textcolor{red}{\scriptsize(+0.40)} & 51.96 \textcolor{red}{\scriptsize(+0.43)} & 45.90 \textcolor{red}{\scriptsize(+0.45)} \\
\checkmark & \checkmark & & &
67.33 \textcolor{red}{\scriptsize(+2.77)} & 66.58 \textcolor{red}{\scriptsize(+3.43)} & 60.63 \textcolor{red}{\scriptsize(+4.69)} & 56.12 \textcolor{red}{\scriptsize(+2.27)} & 54.49 \textcolor{red}{\scriptsize(+2.53)} & 47.97 \textcolor{red}{\scriptsize(+2.07)} \\
\checkmark & \checkmark & \checkmark & &
69.49 \textcolor{red}{\scriptsize(+2.16)} & 68.36 \textcolor{red}{\scriptsize(+1.78)} & 61.76 \textcolor{red}{\scriptsize(+1.13)} & 57.10 \textcolor{red}{\scriptsize(+0.98)} & 55.56 \textcolor{red}{\scriptsize(+1.07)} & 49.41 \textcolor{red}{\scriptsize(+1.44)} \\
\checkmark & \checkmark & \checkmark & \checkmark &
\textbf{71.48} \textcolor{red}{\scriptsize(+1.99)} & \textbf{70.14} \textcolor{red}{\scriptsize(+1.78)} & \textbf{63.87} \textcolor{red}{\scriptsize(+2.11)} &
\textbf{58.64} \textcolor{red}{\scriptsize(+1.54)} & \textbf{57.09} \textcolor{red}{\scriptsize(+1.53)} & \textbf{50.78} \textcolor{red}{\scriptsize(+1.37)} \\
\bottomrule
\end{tabular*}
\end{minipage}
\end{table*}


The complete DCPNet achieves superior prediction accuracy on both datasets. In the image modality, the MAE reaches 0.111 and 0.146 on SI3D-DI and SI3D-DII, respectively. In the LiDAR modality, the MAE reaches 0.157 and 0.167 on the two datasets. The stable accuracy across both modalities indicates that DCPNet can reliably fit the dual-modal occlusion intensity ground truth under single-frame conditions. Since the training and test sets are split by non-overlapping geographical regions, these results also demonstrate that DCPNet generalizes well to unseen areas rather than merely memorizing training-set occlusion patterns.

The ablation results further show that the auxiliary features in both
branches effectively improve prediction accuracy, with consistent trends
across both datasets. For the image branch, introducing the LiDAR point
projection auxiliary feature yields stable improvements across all three
metrics, with
$R^2$
increasing from 0.653 to 0.714 on SI3D-DII. The sparse structural
information from LiDAR projection helps the image branch perceive
occlusion boundaries and reduces semantic ambiguity in heavily occluded
regions. For the LiDAR branch, the improvement from geometric
statistical features is more pronounced, with MAE decreasing from 0.205
to 0.167 on SI3D-DII. This further confirms that geometric
statistical features provide explicit prior guidance for the prediction
network, enabling it to reliably perceive occlusion states from discrete
sparse voxels.

Fig.~\ref{fig:10} presents visual comparison results for both modalities. The
predictions of the complete DCPNet are closer to the ground truth in
occlusion boundary localization and regional distribution, better
recovering the spatial structure of occluded regions, open areas, and
transitional edges. This is consistent with the trends observed in the
quantitative metrics, further validating the positive contribution of
auxiliary features to dual-modal occlusion intensity prediction.

\subsection{Ablation Study of CAMF-Det}
To further analyze the contribution of each component of CAMF-Det, we
conduct two groups of ablation experiments. The first group examines the
impact of DCPNet and OPF on the overall detection performance. The
second group progressively analyzes the contribution of each of the four
OPF sub-modules to the detection performance.

\subsubsection{Overall Method Ablation}
Starting from the single-modal baseline TG-ADet, we progressively
introduce the image branch, DCPNet, and the OPF. The results are shown
in Fig.~\ref{fig:11}.

The multimodal baseline achieves
$\mathrm{mAP}_{\mathrm{BEV}}$
of 65.60\%, 63.85\%, and 56.80\% at the three difficulty levels, with
only limited improvement over TG-ADet. This indicates that
straightforward multimodal fusion cannot effectively handle the spatially varying and modality-dependent degradation caused by ground-object occlusion, and fails to stably
exploit the complementary information from the image modality. Adding
DCPNet to this baseline yields
$\mathrm{mAP}_{\mathrm{BEV}}$
of 64.92\%, 63.26\%, and 55.53\% at the three difficulty levels, showing
no improvement in performance. This suggests that DCPNet, as an
independent prediction module, does not provide positive gains when its
output is not integrated into the detection pipeline, and may even
hinder detection optimization. With the complete CAMF-Det,
$\mathrm{mAP}_{\mathrm{BEV}}$
reaches 71.48\%, 70.14\%, and 63.87\%, improving over the multimodal
baseline by 5.88\%, 6.29\%, and 7.07\%, respectively. This result
demonstrates that the occlusion intensity predicted by DCPNet significantly
improves detection performance once explicitly integrated through OPF.

\subsubsection{OPF Sub-Module Ablation}
This subsection takes TG-ADet as the starting point and progressively
introduces the four OPF sub-modules to evaluate each
component's contribution. Following inter-module
dependencies, the four sub-modules are added in the order of
occlusion-aware response modulation (OPF-Mod), occlusion prior-weighted
fusion (OPF-Fus), occlusion-matched data augmentation (OPF-Aug), and
occlusion-gated contrastive learning (OPF-Con). The results are
presented in Table~\ref{tab:8}.

Introducing detection response modulation on top of TG-ADet improves
$\mathrm{mAP}_{\mathrm{BEV}}$
by 1.19\%, 1.77\%, and 1.50\% at the three difficulty levels. This
module operates on the classification branch of the detection head,
yielding limited but stable gains. Further introducing occlusion
prior-weighted fusion improves
$\mathrm{mAP}_{\mathrm{BEV}}$
by an additional 2.77\%, 3.43\%, and 4.69\%. This module directly
participates in multimodal feature interaction and enables effective
fusion under occlusion prior guidance. Adding dual-modal
occlusion-matched sample generation on this basis further improves the
detection metrics by 2.16\%, 1.78\%, and 1.13\%, indicating that
maintaining occlusion consistency in data distribution through
dual-modal occlusion matching during training provides a notable benefit
to detection performance. Finally, introducing contrastive learning
yields further improvements of 1.99\%, 1.79\%, and 2.10\%. This
demonstrates that conducting contrastive learning after filtering out
low-quality image features can further enhance the image
branch's discriminative capability for fine-grained
differences.

The two groups of ablation results together confirm that the performance
gains of CAMF-Det arise from the synergy between DCPNet and the OPF
sub-module. DCPNet generates dual-modal occlusion priors, while OPF
embeds the outputs of DPM and DCPNet into the detection pipeline through
its four sub-modules, achieving stable improvements under complex
occlusion scenarios.

\section{Conclusion}\label{conclusion}

In this work, we propose CAMF-Det, a closure-aware multimodal fusion framework for LiDAR-camera 3D object detection on UAV platforms, to address the spatially varying and modality-dependent information degradation caused by ground-object occlusion. The proposed method derives dual-modal occlusion intensity from physics-inspired closure modeling, converts it into online predictions, and embeds the resulting priors throughout the detection pipeline, enabling adaptive multimodal detection in occluded UAV scenes. Experimental results on the SI3D-DI and SI3D-DII datasets demonstrate that CAMF-Det outperforms existing single-modal and multimodal methods, with more pronounced advantages on difficult and heavily occluded cases, validating the effectiveness of explicitly modeling and exploiting occlusion priors.

Future work will focus on improving detection accuracy under extreme occlusion, extending the method to more diverse environments and platform configurations, and exploring cross-scene domain adaptation to further enhance the generalization capability of the framework.

\section*{CRediT authorship contribution statement}

\textbf{Yanze Jiang:} Formal analysis, Methodology, Validation, 
Writing -- original draft, Visualization, Software. \textbf{Yanfeng Gu:} Funding acquisition, Supervision, 
Writing -- review \& editing, Conceptualization, Resources. \textbf{Xian Li:} Conceptualization, Investigation, Data curation, 
Supervision, Writing -- review \& editing, Project administration.
\section*{Acknowledgment}

This work was supported by the Major Scientific Instrument Development Program of the National Natural Science Foundation of China (Grant No. 62327803).

\end{document}